# A Physics Prior-Guided Dual-Stream Attention Network for Motion Prediction of Elastic Bragg Breakwaters

Lianzi Jiang[1], Jianxin Zhang[1], Xinyu Han[2,3,*], Huanhe Dong[1], Xiangrong Wang[1]

[1] Shandong University of Science and Technology, College of Mathematics and Systems Science, Qingdao, 266590, China

[2] Shandong University of Science and Technology, College of Civil Engineering and Architecture, Qingdao, 266590, China

[3] Shandong Provincial Key Laboratory of Reliability and Intelligent Protection for Marine Engineering Equipment, Qingdao, 266590, China

* Corresponding author: Xinyu Han (E-mail: hanxinyu_sdust@sdust.edu.cn)

Other authors' institutional emails:

Lianzi Jiang: jianglianzi@sdust.edu.cn

Jianxin Zhang: 202181150008@sdust.edu.cn

Huanhe Dong: donghuanhe@sdkd.net.cn

Xiangrong Wang: skd991278@sdust.edu.cn

## Abstract

Accurate motion response prediction for elastic Bragg breakwaters is critical for their structural safety and operational integrity in marine environments. However, conventional deep learning models often exhibit limited generalization capabilities when presented with unseen sea states. These deficiencies stem from the neglect of natural decay observed in marine systems and inadequate modeling of wave-structure interaction (WSI). To overcome these challenges, this study proposes a novel Physics Prior-Guided Dual-Stream Attention Network (PhysAttnNet). First, the decay bidirectional self-attention (DBSA) module incorporates a learnable temporal decay to assign higher weights to recent states, aiming to emulate the natural decay phenomenon. Meanwhile, the phase differences guided bidirectional cross-attention (PDG-BCA) module explicitly captures the bidirectional interaction and phase relationship between waves and the structure using a cosine-based bias within a bidirectional cross-computation paradigm. These streams are synergistically integrated through a global context fusion (GCF) module. Finally, PhysAttnNet is trained with a hybrid time-frequency loss that jointly minimizes time-domain prediction errors and frequency-domain spectral discrepancies. Comprehensive experiments on wave flume datasets demonstrate that PhysAttnNet significantly outperforms mainstream models. Furthermore, cross-scenario generalization tests validate the model's robustness and adaptability to unseen environments, highlighting its potential as a framework to develop predictive models for complex systems in ocean engineering.

## 1. Introduction

Medium-long period waves, with periods of approximately 10-30 s, are prevalent during cyclonic events such as typhoons and at construction sites along the Maritime Silk Road (Han and Dong, 2023). Characterized by long wavelengths, high wave energy, and slow attenuation, medium-long period waves represent a characteristic condition that must be accounted for in coastal disaster mitigation (Qie et al., 2020). Under such conditions, traditional rigid breakwaters are prone to inducing strong reflection in front of the structure, experience large wave loads, and exhibit high wave transmission (Sammarco et al., 2021; Han et al., 2023; Han and Dong, 2023). Existing studies indicate that, because the characteristic size of a monolithic structure is far smaller than the wavelength of medium-long period waves, effective dissipation is difficult to achieve (Ji et al., 2024). Therefore, to significantly enhance the wave-dissipation performance of breakwaters under medium-long period waves, research into multi-body configurations and novel wave-dissipating mechanisms is required.

In response to these challenges, multi-body configurations have been explored. Among these, multiple breakwaters arranged at regular intervals are referred to as Bragg breakwaters (Mei et al., 1988). Unlike traditional fixed breakwaters, Bragg breakwaters induce Bragg resonance of waves between their structures, thereby impeding wave intrusion (Peng et al., 2019). Physical experiments and theoretical analyses have demonstrated that the number, width, and height of the structures, as well as the seabed slope, significantly affect the Bragg resonance reflection coefficient (Zeng et al., 2017; Ning et al., 2022; Xie et al., 2023). However, when the number of Bragg breakwaters is three, the induced resonance reflection coefficient reaches only about 0.62 (Ouyang et al., 2016). The structure exhibits significant wave attenuation only near the resonance frequency, while incident waves of other frequencies are more readily transmitted.

To address these limitations, an elastic Bragg breakwater is proposed to synergistically activate three wave-dissipation mechanisms: (1) increase the effective breakwater height and the deformation of the elastic body, thereby inducing wave breaking and enhancing energy dissipation (Lan et al., 2010); (2) absorb wave energy via elastic deformation, reducing wave impact and reflection; and (3) induce Bragg resonance and, by tuning the resonant frequency and bandwidth, achieve optimal disaster-mitigation performance, thereby substantially improving dissipation under medium-long-period waves. Once elasticity is introduced, however, the breaking threshold, turbulence and vertex characteristics, and formation mechanism of the resonant band are all influenced by structural dynamics, and therefore conclusions derived for rigid systems no longer apply directly. Moreover, theoretical and numerical predictions become highly parameter-sensitive and strongly dependent on experimental validation.

Under medium-long-period wave action, elastic Bragg breakwaters are subjected to large deformations and internal forces, which entail risks of fatigue, buckling, and even structural failure (Hascoët and Jacques, 2025). To calibrate design parameters, stabilize the resonant band, and provide a consistent baseline for operational monitoring, it is essential to establish reliable structural-response prediction capabilities at the design stage. Accurate prediction is also a prerequisite for real-time assessment of wave-dissipation performance and identification of extreme sea states (Zhang et al., 2025). However, because wave–elastic coupling is complex and traditional theoretical and numerical models are computationally intensive, they cannot readily satisfy rapid and real-time requirements. Conventional approaches for computing the structural response of marine engineering systems are broadly classified into potential-flow theory and viscous-flow theory. The former necessitates the introduction of prescribed hydrodynamic coefficients, which inherently constrains its applicability in regimes characterized by pronounced nonlinearity or excessive motion amplitudes (Bharath et al., 2018). The latter, while capable of capturing intricate phenomena such as wave breaking and vortex evolution, demands exceedingly fine spatial discretization and small temporal steps, resulting in prohibitive computational costs (Landesman et al., 2024). Consequently, neither framework is well-suited for rapid yet accurate prediction of structural responses.

Machine learning (ML) techniques, by virtue of their strong regression capabilities, can effectively characterize nonlinear interactions among multiple variables. In particular, deep learning models are able to assimilate extensive input–output mappings during the training phase, enabling near-instantaneous inference at speeds vastly surpassing those of numerical simulations (Juan et al., 2025). This renders them particularly amenable to real-time monitoring and rapid design iteration. A data-driven paradigm, when used in lieu of computationally intensive, fully physics-based modeling, can retain predictive accuracy while dramatically accelerating computations and improving adaptability to complex nonlinearities and uncertainties (Wang et al., 2024). Such an approach is especially well-suited to elastic Bragg breakwaters, where strong fluid-structure coupling and stringent real-time constraints prevail—most notably in applications involving design-parameter optimization, online condition monitoring and early-warning systems, and rapid assessment of extreme sea states. In the realm of time-series prediction, traditional ML methods—such as support vector regression (Zhu et al., 2019), random forests (Wang et al.,2022), and gradient-boosted trees (Han and Dong, 2025)—can achieve moderate accuracy under certain conditions. However, they exhibit clear limitations when applied to structural-response forecasting in oceanic environments. First, these algorithms generally depend on manually engineered features, which are often inadequate for fully capturing the intricate temporal patterns embedded in wave–structure interactions, and lack the capacity for end-to-end representation learning (Zhong et al., 2016). Second, they typically operate on fixed-length historical windows, rendering them ill-equipped to model the long-term dependencies that are prevalent in structural responses (Chen et al., 2023). Third, when confronted with nonstationary, strongly nonlinear, and inherently multiscale marine dynamic processes, such models often require extensive preprocessing and simplifying assumptions (Tian et al., 2025), thereby constraining their predictive performance. Moreover, field monitoring data are frequently marred by noise and missing values, to which these methods exhibit high sensitivity, resulting in insufficient robustness.

These shortcomings limit the applicability of conventional ML techniques to the prediction of structural responses in strongly coupled systems with pronounced long-period effects. This motivates the adoption of deep learning architectures, which can autonomously extract salient features, capture long-range dependencies, and perform end-to-end learning, thereby enhancing both predictive accuracy and generalization capability. Recurrent Neural Networks (RNNs) and their advanced variant, Long Short-Term Memory (LSTM) (Hochreiter and Schmidhuber, 1997), have emerged as prominent solutions for time series forecasting tasks. LSTMs, specifically designed to address the vanishing and exploding gradient problems inherent in simple RNNs, enable effective learning of long-range dependencies. In ocean engineering applications, LSTM has demonstrated widespread utility in predicting ship motion, semi-submersible dynamics, mooring tension, and floating platform responses, consistently outperforming traditional methods in capturing short-term nonlinear dynamics (Peng et al., 2019; Neshat et al., 2021; Miyanawala et al., 2024; Hu et al., 2025). Several studies underscore LSTM's effectiveness: Saad et al. (2021) demonstrated LSTM's superior performance over Multilayer Perceptrons (MLP) in motion prediction for mooring line failure detection. The benefits of multivariate inputs have been extensively documented. Guo et al. (2021) successfully integrated motion data and wave information within an LSTM framework for real-time prediction of semi-submersible heave and surge motions. Similarly, Shi et al. (2023) introduced a Multi-Input LSTM (MI-LSTM), showing enhanced prediction accuracy for floating offshore wind turbine (FOWT) platforms by incorporating wave elevation and mooring forces alongside platform motion response, surpassing single-input LSTM and Convolutional Neural Network (CNN) models. Additionally, Yin et al. (2025) achieved reliable dynamic response predictions for semi-submersible platforms using Bi-directional LSTM (Bi-LSTM) with optimized hyperparameters, leveraging both wave conditions and historical motion data. However, LSTM models face inherent architectural limitations, including computational inefficiencies from sequential processing and challenges in modeling long-range dependencies.

To address the inherent limitations of LSTM architectures, researchers have predominantly focused on developing hybrid prediction models that employ a "decompose-and-ensemble" framework to leverage the complementary strengths of specialized modules. A key strategy within this framework incorporates CNN or attention mechanisms to enhance feature extraction capabilities. For instance, Xiang et al. (2022) developed SATCN-LSTM, combining a self-attention mechanism with a Temporal Convolutional Network (TCN) and LSTM to extract crucial temporal features for ultra-short-term wind power forecasting, demonstrating superior performance over multiple benchmarks. In a similar vein, Xu and Ji (2024) developed a composite model for short-term motion prediction of semi-submersible platforms, demonstrating that a one-dimensional CNN-BiLSTM architecture with attention achieved optimal predictive accuracy. Tang et al. (2025) further enhanced a CNN-LSTM-ATT model using Chebyshev polynomials for mooring line tension forecasting, achieving significant improvements in accuracy. The optimization of these complex architectures has become increasingly important, with studies by Zhang et al. (2023) and Abou Houran et al. (2023) implementing advanced algorithms for hyperparameter tuning, yielding substantial performance improvements over traditional approaches. Parallel to these developments, researchers have explored the integration of signal processing techniques, including Empirical Mode Decomposition (EMD) and its advanced variants such as EEMD (Wu et al., 2009) and ICEEMDAN (Colominas et al., 2014), to address non-stationary signal challenges. This methodology involves decomposing complex time series into simpler Intrinsic Mode Functions (IMFs), independently forecasting these components, and then combining them (Zhang et al., 2020; Hao et al., 2022). Applications of this approach include floating platform motion forecasting using EEMD with Support Vector Regression (Hong et al., 2019) and multivariate significant wave height prediction through ICEEMDAN-LSTM integration (Pang and Dong, 2023). Despite the notable performance improvements achieved by these hybrid approaches, fundamental limitations persist. Most critically, models based on LSTM architectures inherit the inherent constraints of recurrent neural networks: limited capacity for capturing long-range dependencies, vulnerability to error accumulation, and reduced computational efficiency due to sequential processing requirements. These limitations underscore the need for a unified architecture capable of comprehensively modeling complex temporal dependencies without such constraints.

The Transformer architecture (Vaswani et al., 2017) has emerged as a compelling solution for ocean forecasting applications. Through its self-attention mechanism, it efficiently processes sequences in parallel and models long-range dependencies between any two positions, thereby overcoming the sequential limitations inherent in RNNs. Recent investigations have demonstrated the efficacy of Transformer-based approaches across diverse marine and offshore engineering domains. Triviño et al. (2023) implemented a Transformer-based framework for structural damage detection in offshore wind turbine systems, achieving superior accuracy and computational efficiency compared to conventional methods. Similarly, Pokhrel et al. (2022) developed a Transformer model to predict residual errors between numerical weather predictions and observational data for significant wave height forecasting, substantially outperforming existing predictive methodologies. Tan et al. (2025) introduced a Swin-Transformer architecture for regional wave field prediction, demonstrating enhanced capability in capturing complex spatio-temporal dependencies for short-term forecasting horizons. Furthermore, Yuan (2024) empirically validated that an enhanced Transformer architecture exhibited superior performance relative to LSTM networks in predicting mooring tension dynamics of semi-submersible platforms.

However, the direct application of existing Transformer-based models to the motion response prediction of elastic Bragg breakwaters is not well-suited, presenting several key challenges:

1. **Neglect of recent-history dependencies**: Purely data-driven self-attention models often fail to capture the temporal decay characteristics inherent in physical systems, where the current motion response state exhibits stronger correlations with recent historical data than with temporally distant states. This limitation makes the model susceptible to establishing spurious correlations that violate underlying physical principles.

2. **Inadequate modeling of wave-structure interaction (WSI)**: The mechanism underlying WSI in elastic Bragg breakwaters involves a bidirectional process wherein structural motion response and the surrounding wave excitation field mutually influence each other. In practical monitoring systems, phase differences in the recorded time series, caused by the spatial separation between wave sensors and the structure, establish the wave-structure phase relationship as a governing mechanism. However, conventional attention architectures suffer from two limitations in capturing this interaction: firstly, their unidirectional information flow is incapable of representing the reciprocal coupling between wave excitation and structural response; secondly, their standard similarity metrics assign attention weights solely based on temporal feature similarity, lacking the inductive bias necessary to recognize phase relationships.

3. **Time-domain-only supervision**: Supervision exclusively in the time domain overlooks the inherent frequency characteristics of motion responses. Although it may achieve high numerical accuracy, this approach inaccurately represents the true dynamical properties of the motion responses.

To address these limitations, this study proposes a novel Physics Prior-Guided Dual-Stream Attention Network (PhysAttnNet). The core idea of this network lies in decoupling the motion prediction into two parallel and complementary attention streams to separately capture the internal dynamic evolution of the structure and the external wave-structure coupling feedback. Specifically, we design dual-stream attention architecture. The decay bidirectional self-attention (DBSA) module is designed to model motion responses by upweighting the influence of more recent states through temporal decay used as an inductive bias. Concurrently, the phase differences guided bidirectional cross-attention (PDG-BCA) module employs a cosine-based bias within a bidirectional cross-computation paradigm. This approach directly captures bidirectional WSI and phase relationships, enabling the physically consistent simulation of the wave-structure coupling process. To achieve deep integration of these two types of information, we introduce a global context fusion (GCF) module. This module utilizes the output of PDG-BCA to generate a global summary, which is then interactively fused with the features from DBSA. At the model optimization level, we design a hybrid time-frequency loss function that simultaneously supervises time-domain errors and frequency-domain spectral errors, guiding the network to learn motion responses with both numerical accuracy and spectral fidelity. The contributions of this study are as follows:

1. We propose PhysAttnNet, a novel architecture that decouples the complex motion prediction task into two specialized streams: the internal temporal evolution and the external WSI. These streams are synergistically integrated by a GCF module, enabling deep and adaptive feature interaction guided by physical priors.
2. We develop a dual-stream attention framework. The DBSA mechanism incorporates a learnable decay bias to emphasize recent-history dependencies, whereas the PDG-BCA mechanism employs a cosine-based bias within a bidirectional cross-computation paradigm to explicitly model WSI and phase relationships, enhancing prediction robustness.
3. We introduce a hybrid time-frequency loss function that jointly optimizes prediction accuracy in the time domain and spectral fidelity in the frequency domain, ensuring that the model produces numerically accurate while exhibiting high spectral fidelity.
4. We conduct comprehensive experiments on wave flume datasets under diverse conditions. The results demonstrate that the proposed PhysAttnNet significantly outperforms mainstream models in both prediction accuracy and generalization capability.

The remainder of this study is structured as follows: Section 2 details the proposed PhysAttnNet framework. Section 3 describes the dataset construction methodology, including physical tests and numerical simulations. Section 4 outlines the experimental setup, covering evaluation metrics, baseline models, and hyperparameter configurations. Section 5 presents the experimental results, including overall performance, analysis across varying prediction horizons, ablation studies, and generalization evaluation under diverse wave conditions. Section 6 concludes the study with key findings, limitations, and directions for future research.

## 2 Model

This section offers a comprehensive description of the designed PhysAttnNet model, covering its overall framework and technical specifics. As illustrated in Fig. 1, the model is architected around a dual-stream attention backbone. The first stream employs the DBSA mechanism, which incorporates the physical prior of temporal decay by introducing a learnable bias to assign higher weight to recent historical information. Concurrently, the second stream utilizes the PDG-BCA module, specifically designed to model the WSI through a bidirectional computational paradigm and a phase-biased attention mechanism. To enable synergistic integration of the two streams, the GCF module generates a global context vector from the output of PDG-BCA and adaptively fuses it with the representations from DBSA. Finally, the model's training is supervised by a hybrid time-frequency loss function.

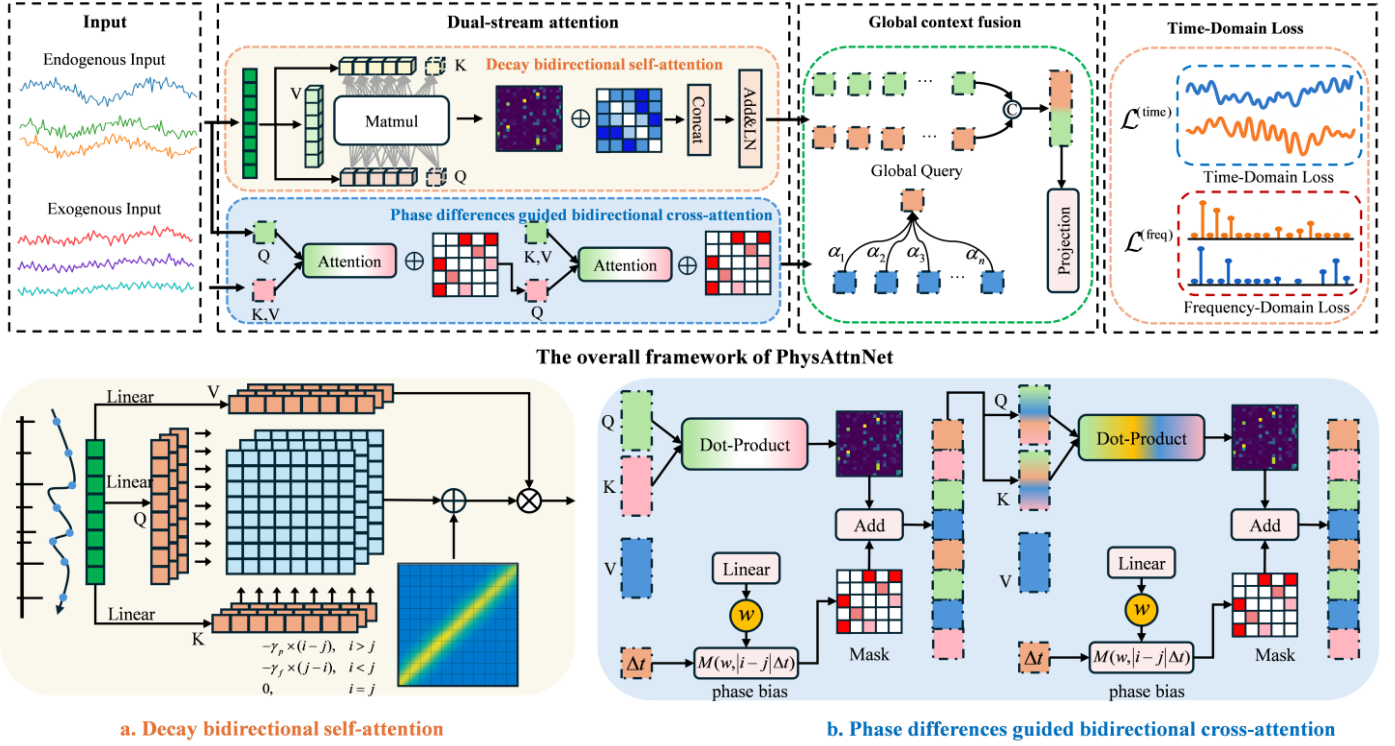

Fig. 1. The overall framework of PhysAttnNet.

## 2.1. Decay bidirectional self-attention

In conventional self-attention mechanisms, the allocation of attention weights is primarily determined by feature similarity. This approach shows clear limitations when dealing with temporal signals: even if two time steps are far apart, the model may still assign high attention weights to them as long as their waveform features appear similar. However, for predicting the structural response time series of elastic Bragg breakwaters, such similarity-based attention allocation does not align with physical reality. In fact, the temporal nature of structural responses dictates that the correlation between the current motion response state and distant historical gradually weakens as the time interval increases. Ignoring this attenuation effect leads the model to overestimate the contribution of distant states, underestimate the extreme response amplitudes, and ultimately reduce overall predictive accuracy.

To solve this problem, we have constructed an attenuated bidirectional sub-attention mechanism. The dynamics of waves and elastic Bragg breakwaters interaction systems is fundamentally governed by structural dynamics principles where damping constitutes a critical parameter. This dynamic behavior can be mathematically described through an exponential decay factor of the form $e^{-\zeta \omega_n (t-\tau)}$, where $\xi$ represents the effective decay ratio and $\omega_n$ denotes the natural frequency (Chopra et al., 2007). This exponential attenuation embodies the physical principle that historical influences diminish naturally with increasing temporal separation from the current time step. However, existing self-attention mechanisms fail to account for this temporal decay characteristic, as weight assignment depends exclusively on feature similarity, potentially leading to overemphasis on temporally distant events. To address this, we propose DBSA, a novel mechanism designed to explicitly incorporate the physical prior of temporal decay into its attention computation, as illustrated in Fig. 1 (a).

Let $X \in \mathbb{R}^{L \times d_{\text{model}}}$ denote the input sequence, where $L$ is the sequence length and $d_{\text{model}}$ is the embedding dimension. For head $h \in \{1, \ldots, H\}$, the input sequence $X$ is linearly projected into query ($Q_h$), key ($K_h$), and value ($V_h$) matrices:

$$Q_h = XW_Q^{(h)}, K_h = XW_K^{(h)}, V_h = XW_V^{(h)}, \tag{1}$$

where $W_Q^{(h)} \in \mathbb{R}^{d_{\text{model}} \times d}$, $W_K^{(h)} \in \mathbb{R}^{d_{\text{model}} \times d}$, $W_V^{(h)} \in \mathbb{R}^{d_{\text{model}} \times d}$ represent the learnable matrices and $d$ is the feature dimension. For head $h$, the attention weight $\alpha^{(h)}(i,j)$ quantifying the dependency between the vector $q_i^{(h)}$ and the vector $k_j^{(h)}$ is computed by applying the softmax function:

$$\alpha^{(h)}(i,j) = \text{softmax}\left(q_i^{(h)} k_j^{(h)T}\right),$$
$$= \frac{\exp\left(q_i^{(h)} k_j^{(h)T}\right)}{\sum_{l}^{L} \exp\left(q_i^{(h)} k_l^{(h)T}\right)}. \quad (2)$$

Then, we discretize the continuous time difference $t - \tau$ into $|i - j|\Delta t$. $\Delta t = 0.05$ represents the constant sampling interval of the time series data. Crucially, we recognize that the decay $\xi \omega_n$ is not a constant value. Directly calculating this value is extremely difficult and inaccurate. Therefore, we use a learnable parameter $\gamma$ to replace the decay. To incorporate the decay factor $e^{-\gamma |i-j|\Delta t}$ into the attention calculation, we leverage the mathematical properties of the softmax function. Adding a bias term $C$ to a specific input element of the softmax is equivalent to multiplying the corresponding unnormalized exponential by $e^C$ before normalization. Therefore, we only need to add its logarithm, which is $-\gamma |i-j|\Delta t$, to the original attention score before the softmax computation. The decay bias is parameterized as a function of the relative temporal distance $|i - j|$:

$$D^{(h)}(i,j) = \begin{cases} -\gamma_p \cdot (i-j) \cdot \Delta t & \text{if } i > j \\ 0 & \text{if } i = j, \\ -\gamma_f \cdot (j-i) \cdot \Delta t & \text{if } i < j \end{cases} \quad (3)$$

where $\gamma_p$ and $\gamma_f$ are learnable non-negative parameters representing temporal decay rates. The model learns these decay rates directly from data, enabling adaptive adjustment of temporal dependency modeling according to different sea states and structural configurations, thereby enhancing predictive robustness and generalization capability.

The final attention scores $\tilde{\alpha}^{(h)}(i,j)$ are then computed by incorporating this bias:

$$\tilde{\alpha}^{(h)}(i,j) = \frac{\exp\left(q_i^{(h)} k_j^{(h)T} + D^{(h)}(i,j)\right)}{\sum_{l}^{L} \exp\left(q_i^{(h)} k_l^{(h)T} + D^{(h)}(i,l)\right)}. \quad (4)$$

When the feature similarity term $q_i^{(h)} k_j^{(h)T}$ is neglected, the attention scores are simplified to:

$$\tilde{\alpha}^{(h)}(i,j) \propto \exp\left(D^{(h)}(i,j)\right). \quad (5)$$

The output for position $i$ in head $h$ is then computed by performing a weighted sum of the value vectors $v_j^{(h)}$ in the value matrix $V_h$:

$$\text{head}_h(i) = \sum_{j=1}^{L} \tilde{\alpha}^{(h)}(i,j) v_j^{(h)}. \quad (6)$$

To capture information from different representation subspaces, the outputs of all attention heads are concatenated along the feature dimension. This concatenated matrix is then passed through a final linear projection layer, governed by a weight matrix $W^O \in \mathbb{R}^{Hd \times d_{\text{model}}}$, to produce the final output of the multi-head attention block:

$$\text{output}^{DBSA} = \text{Multihead}(Q, K, V) = \text{Concat}(\text{head}_1, \ldots, \text{head}_H) W^O, \quad (7)$$

where $\text{output}^{DBSA} \in \mathbb{R}^{L \times d_{\text{model}}}$ represents the final output of the DBSA. This process enables each attention head $h$ to dynamically balance feature-driven attention with temporal decay dynamics, thereby yielding a more robust sequence representation.

## 2.2 Phase differences guided bidirectional cross-attention

In the monitoring system, the spatial separation between the wave surface measurement points and the elastic Bragg breakwater introduces phase differences in the recorded time series. When the incident wave frequency approaches the structural natural frequency or the Bragg resonance frequency, the phase relationship becomes a key factor determining whether the structural response amplitude will be amplified. The standard cross-attention mechanism allocates weights solely based on feature similarity. It cannot recognize or emphasize the phase relationship between the wave and the structure, which may cause time series with different phase differences to be misinterpreted as strongly correlated, resulting in distorted predictions. To address this, we introduce a phase bias term into the attention score, allowing the attention weights to depend not only on feature similarity but also on the modulation of phase differences. The PDG-BCA module explicitly models the WSI dynamics through two components: 1) a phase-biased attention mechanism that modulates attention weights based on phase-coherence, enabling the model to selectively focus on phase difference interactions, and 2) a bidirectional cross-computation paradigm that computes the mutual influence between wave and motion response time-series.

### 2.2.1 Phase-biased attention

To integrate the phase relationship prior into the cross-attention mechanism, a framework for characterizing temporal phase coherence between time points is required. Given two time points $t_i$ and $t_j$ sampled from a periodic signal of frequency $f$, their phase difference is expressed as:

$$\Delta\phi = 2\pi f\left(t_i - t_j\right). \tag{8}$$

Consequently, given the difference between query at position $i$ and the key at position $j$, and a learnable frequency parameter $w$, the phase bias term is defined as:

$$M\left(w, |i-j|\Delta t\right) = \cos\left(2\pi w|i-j|\Delta t\right), \tag{9}$$

where $\Delta t = 0.05$ represents the constant sampling interval of the time series data. The range of this function is $[-1, 1]$, and its values are entirely determined by the phase $w|i-j|\Delta t$. When the $w$ in the model learns and converges to the system's resonance frequency $w_r$, the value of $M(w_r, |i-j|\Delta t)$ will reflect the physical phase relationship between the wave excitation and the structural response. We can formalize this as follows:

$$\text{if } w|i-j|\Delta t = n, \ n \in \mathbb{Z} \implies M\left(w, |i-j|\Delta t\right) = \cos(2\pi n) = +1, \tag{10}$$

$$\text{if } w|i-j|\Delta t = n+1/4, \ n \in \mathbb{Z} \implies M\left(w, |i-j|\Delta t\right) = \cos(2\pi(n+1/4)) = 0, \tag{11}$$

$$\text{if } w|i-j|\Delta t = n+1/2, \ n \in \mathbb{Z} \implies M\left(w, |i-j|\Delta t\right) = \cos(2\pi(n+1/2)) = -1. \tag{12}$$

Specifically, when $w|i-j|\Delta t = n$, $\cos(2\pi w|i-j|\Delta t) = 1$, when $w|i-j|\Delta t = n+1/4$, $\cos(2\pi w|i-j|\Delta t) = 0$, and when $w|i-j|\Delta t = n+1/2$, $\cos(2\pi w|i-j|\Delta t) = -1$. Crucially, $w$ is defined as a learnable parameter specific to each attention head rather than a predefined constant. This design transforms each attention head into an adaptive filter that autonomously identifies relevant resonant frequencies from training data, thereby enforcing physics-informed feature extraction.

The process begins with the standard linear projection of the input sequence $X \in \mathbb{R}^{L \times d_{\text{model}}}$ into query ($Q_h$), key ($K_h$) and value ($V_h$) matrices for each attention head $h \in \{1, \ldots, H\}$:

$$Q_h = XW_Q^{(h)}, K_h = XW_K^{(h)}, V_h = XW_V^{(h)}, \tag{13}$$

where $W_Q^{(h)} \in \mathbb{R}^{d_{\text{model}} \times d}, W_K^{(h)} \in \mathbb{R}^{d_{\text{model}} \times d}, W_V^{(h)} \in \mathbb{R}^{d_{\text{model}} \times d}$ represent weight matrices for head $h$. The phase bias term is defined as:

$$B_{ij}^{(h)} = M^{(h)}\left(w, |i-j| \Delta t\right). \tag{14}$$

The modified attention score between a query $q_i^{(h)}$ and a key $k_j^{(h)}$ is formulated as:

$$\alpha^{(h)}(i,j) = \frac{\exp\left(q_i^{(h)} k_j^{(h)T} + B_{ij}^{(h)}\right)}{\sum_{l}^{L} \exp\left(q_i^{(h)} k_l^{(h)T} + B_{il}^{(h)}\right)}, \tag{15}$$

and the output for position $i$ in head $h$ is:

$$head_h(i) = \sum_{j=1}^{L} \alpha^{(h)}(i,j) v_j^{(h)}, \tag{16}$$

where $L$ is the sequence length. To allow the model to jointly attend to information from different representation subspaces and at different phase frequencies, the outputs of all attention heads are concatenated:

$$Multihead(Q, K, V) = Concat(head_1, \ldots, head_H) W^O, \tag{17}$$

where $W^O \in \mathbb{R}^{Hd \times d_{model}}$ is the weight matrix. This mechanism serves as a phase-coherence-driven attention modulator, adaptively reweighting conventional cross-attention scores to emphasize interactions with strong temporal-spectral alignment.

**2.2.2 Bidirectional computational paradigm**

To capture the bidirectional nature of WSI, we employ a sequential two-stage computational process. Let $X_{ext} \in \mathbb{R}^{L \times d_{model}}$ denote the external wave subsequence, $X_{int} \in \mathbb{R}^{L \times d_{model}}$ denote the internal response subsequence, where $L$ is the sequence length and $d_{model}$ is the feature dimension. The initial stage models the direct impact of wave excitation on the structural motion. In this stage, the internal response $X_{int}$ acts as the query, the external wave $X_{ext}$ serves as both key and value:

$$Q_{int} = X_{int} W_1^Q, K_{ext} = X_{ext} W_1^K, V_{ext} = X_{ext} W_1^V, \tag{18}$$

$$X_{int}^{(1)} = Multihead(Q_{int}, K_{ext}, V_{ext}), \tag{19}$$

where $W_1^Q \in \mathbb{R}^{d_{model} \times d}$, $W_1^K \in \mathbb{R}^{d_{model} \times d}$, $W_1^V \in \mathbb{R}^{d_{model} \times d}$ represent weight matrices and $d$ is the feature dimension. $Multihead(\cdot)$ denotes the phase-biased attention mechanism. The second stage builds upon the first to model the feedback mechanism. The external wave state $X_{ext}$ now acts as the query, $X_{int}^{(1)}$ serves as both key and value:

$$Q_{ext} = X_{ext} W_2^Q, K_{int} = X_{int}^{(1)} W_2^K, V_{int} = X_{int}^{(1)} W_2^V \tag{20}$$

$$output^{RI-BCA} = Multihead(Q_{ext}, K_{int}, V_{int}) \tag{21}$$

where $W_2^Q \in \mathbb{R}^{d_{model} \times d}$, $W_2^K \in \mathbb{R}^{d_{model} \times d}$ and $W_2^V \in \mathbb{R}^{d_{model} \times d}$ are weight matrices. $Multihead(\cdot)$ denotes the phase-biased attention mechanism. This updated representation, $output^{RI-BCA} \in \mathbb{R}^{L \times d_{model}}$, thus captures the bidirectional nature of WSI by reflecting the structure's influence on its surrounding wave environment.

This two-stage process forms the structural backbone of PDG-BCA module, establishing a computational framework that explicitly captures the inherent action–reaction dynamics of WSI. The key innovation lies in the integration of the phase-biased mechanism into every cross-attention computation. By synergistically combining macroscopic bidirectional information flow with microscopic attention biasing, the PDG-BCA module enables comprehensive modeling of complex WSI dynamics.

**2.3 Global context fusion**

To effectively fuse the intra-variable dynamics captured by DBSA and the inter-variable causal influences modeled by PDG-BCA, we propose the GCF module, as illustrated in the global context fusion section of Fig. 1. First, to generate the global context vector, we employ an attention-based pooling mechanism. A learnable global query vector, $q^{global} \in R^{1 \times d_{model}}$, is introduced to attend to the full output sequence from the PDG-BCA module, $output^{RI-BCA} \in R^{L \times d_{model}}$, which serves as both the key and value. This operation is formulated as:

$$s = Attention\left(q^{global}, output^{RI-BCA}, output^{RI-BCA}\right) \tag{22}$$

where $s \in R^{1 \times d_{model}}$ summarizes the most salient aspects of the inter-variable causal influences. In the second stage, this global context vector $s$ is integrated with the DBSA output representation. To achieve this, we first broadcast $s$ across the time dimension $L$ to create $s_{broadcast} \in R^{L \times d_{model}}$. This expanded context is then concatenated with the DBSA output $output^{DBSA} \in R^{L \times d_{model}}$ along the feature dimension. Finally, the final prediction $Y^{pred}$ is formulated as:

$$Y^{pred} = Linear\left(Concat\left(output^{DBSA}, s_{broadcast}\right)\right) \tag{23}$$

## 2.4 Hybrid time-frequency loss function

To address the insensitivity of conventional point-wise loss functions to the spectral characteristics of time series, we formulate a hybrid time-frequency loss, as illustrated in the Time-frequency loss section of Fig. 1. The total loss, $\mathcal{L}_{total}$, is a composite of a principal time-domain loss, $\mathcal{L}_{time}$, and an auxiliary frequency-domain loss $\mathcal{L}_{freq}$.

The time-domain loss, $\mathcal{L}_{time}$, is the Mean Absolute Error (*MAE*) between the predicted sequence $Y^{pred}$ and the ground truth $Y^{true}$, ensuring point-wise accuracy:

$$\mathcal{L}_{time} = \frac{1}{N} \sum_{i=1}^{N} \left|Y_i^{pred} - Y_i^{true}\right| \tag{24}$$

where $N$ denotes the prediction time step, $Y_i^{pred}$ represents the predicted value at time step $i$, and $Y_i^{true}$ represents the measured value at time step $i$.

The frequency-domain loss, $\mathcal{L}_{freq}$, is designed to penalize discrepancies within the spectral domain. Specifically, the predicted sequence and the ground truth are first transformed into the frequency domain using the Fast Fourier Transform (FFT), resulting in their corresponding complex-valued frequency representations $\bar{Y}^{pred}$ and $\bar{Y}^{true}$. The frequency-domain loss is then computed as the *MAE* between these two complex-valued spectra:

$$\bar{Y}^{pred} = F\left(Y^{pred}\right), \quad \bar{Y}^{true} = F\left(Y^{true}\right) \tag{25}$$

$$\mathcal{L}_{freq} = \frac{1}{N} \sum_{i=1}^{N} \left(\frac{1}{K} \sum_{a=1}^{K} \left|\bar{Y}_{i,a}^{pred} - \bar{Y}_{i,a}^{true}\right|\right) \tag{26}$$

where $K$ is the number of frequency bins, $\bar{Y}_{i,a}^{pred}$ denotes the a-th frequency component of the predicted value at time step $i$, and $\bar{Y}_{i,a}^{true}$ denotes the a-th frequency component of the measured value at time step $i$. This approach ensures that differences in spectral characteristics are effectively captured and penalized, improving the model's ability to accurately represent spectral information.

The final hybrid loss $\mathcal{L}_{total}$ is a weighted sum of these two components:

$$\mathcal{L}_{total} = (1 - \lambda)\mathcal{L}_{time} + \lambda \mathcal{L}_{freq} \tag{27}$$

where $\lambda$ is a hyperparameter that balances the importance between time-domain accuracy and frequency-domain spectral fidelity. This hybrid loss encourages the model's forecasts achieve both numerical precision and physically plausible.

# 3. Dataset Setup

The main focus of this study is to propose a prediction method for the structural response of elastic Bragg breakwaters. Considering the scarcity and difficulty of obtaining data in practical engineering, the dataset in this work is primarily constructed through physical tests and numerical simulations. A numerical wave flume is established using SPH, and its accuracy is validated against physical model tests. Based on validated numerical model, extensive simulations are carried out to generate the datasets required for neural network.

## 3.1 Physical model

Before hydrodynamic tests, it is necessary to determine the mechanical properties of the elastomer material. The stress-strain curve was measured by an electronic universal testing machine. The elastic modulus of the polyurethane elastomer is found to be 30.91Mpa, and the Poisson ratio is 0.27. Laboratory tests were conducted in a 30 m long, 1 m wide, and 1.5 m deep wave flume at Shandong Provincial Key Laboratory of Reliability and Intelligent Protection for Marine Engineering Equipment. The wave generation system consisted of a piston-type wave paddle with a maximum stoke length of 1.8 m. A slope sponger layer was set on the ends of the flume to absorb waves and reduce reflection. The arrangement of the vertical elastic breakwater is shown in Fig. 2. The physical model setup is show in Fig. 3. The wave condition is regular waves. The water depth is 0.8 m. The incident wave height and period are 0.06 m and 2.19 s. The height and width of vertical elastic breakwater are 0.7 m and 0.06 m, respectively. there are two wave gauges were set in front of the VFB with an interval of 1 m. Other two wave gauges were arranged behind the VFB with an interval of 0.5 m. The layout is exhibited in Fig. 1. The digital image correlation (DIC) is used to measure the deformation of VFB section.

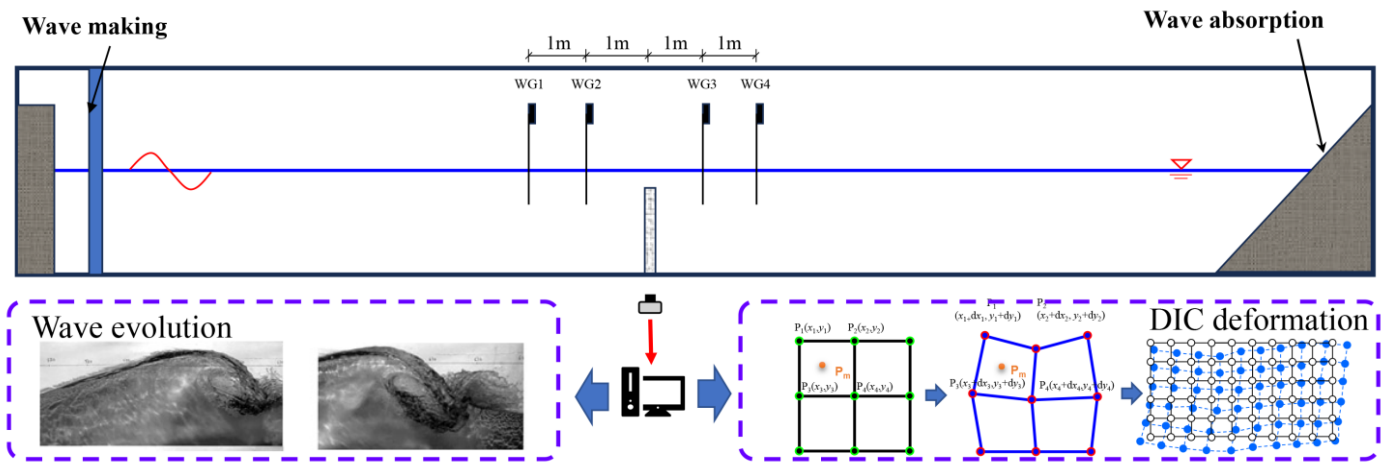

Fig. 2. Physical model schematic of waves and vertical elastic breakwater interaction

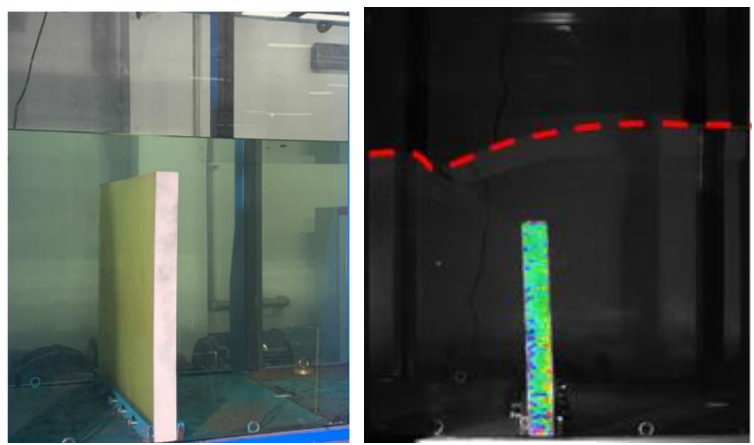

Fig. 3. Physical model setup

## 3.2 Numerical model

The foundations of the SPH method are based on local convolution of neighbouring particles. Introducing a kernel weight function $W$, and discretizing for the particle system. The mass and momentum equations that govern the dynamics of fluid motion are given in Lagrangian particle discrete from by:

$$\frac{D\rho_a}{Dt} = \sum_b m_b (\mathbf{v}_a - \mathbf{v}_b) \bullet \nabla_a W_{ab} + 2\delta h c_0 \sum_b m_b \left(\frac{\rho_a}{\rho_b} - 1\right) \frac{\mathbf{x}_{ab}}{x_{ab}^2 + \eta^2} \bullet \nabla_a W_{ab} \quad (28)$$

$$\frac{D\mathbf{v}_a}{Dt} = -\sum_b m_b \left(\frac{p_a + p_b}{\rho_a \rho_b}\right) \nabla_a W_{ab} + \sum_b m_b \left(\frac{4\upsilon \mathbf{x}_{ab} \bullet \nabla_a W_{ab}}{(\rho_a + \rho_b)(x_{ab}^2 + \eta^2)}\right) \mathbf{v}_{ab} + \sum_b m_b \left(\frac{\tau_{ij}^b}{\rho_b^2} + \frac{\tau_{ij}^a}{\rho_a^2}\right) \nabla_a W_{ab} + \mathbf{g}_a \quad (29)$$

where $\rho$, $t$, $m$, $p$, $\upsilon$ are density, time, mass, pressure, and kinematic viscosity, respectively. $\mathbf{x}$, $\mathbf{v}$, $\mathbf{g}$ are the spatial coordinates, velocity and gravity, respectively. $\mathbf{x}_{ab} = \mathbf{x}_a - \mathbf{x}_b$ represents the distance between particle $a$ and $b$. $x_{ab} = |\mathbf{x}_{ab}|$, $\mathbf{v}_{ab} = \mathbf{v}_a - \mathbf{v}_b$, and $\nabla_a W_{ab} = \nabla_a W(\mathbf{x}_a - \mathbf{x}_b, h)$. $h$ is smooth length of particle support domain. The second term on the right-hand of Eq. (19) is the density diffusion term, which serves to suppress large pressure oscillation. The parameter $\delta$ determines the intensity of this diffusion term and needs to be tuned according to the specific problem. $c_0$ is the initial sound speed. $\eta = 0.1h$ prevents the emergence of singularities when $x_{ab}$ approaches zero. The second term on right hand of Eq. (29) is laminar viscous stress. The sub-particle scale model, third term on right hand of Eq. (29), is used to describe the effect of turbulence. $\tau_{ij}$ is sub-particle stress tensor.

In this SPH framework, the fluid is assumed to be a weakly compressible fluid. The fluid pressure id determined by the equation of state based on the particle density. The compressibility of the fluid can be regulated to the speed of sound. The equation of state as follow:

$$p = \frac{c_0^2 \rho_0}{\gamma}\left[\left(\frac{\rho}{\rho_0}\right)^\gamma - 1\right] \quad (30)$$

where $\gamma=7$, $\rho_0=1000 kg/m^3$. To ensure that the change in fluid density is less than 0.1%, the reference sound of speed $c_0$ is at least 10 times the maximum fluid velocity.

Under the Lagrangian framework, the governing equations of the elastomer structure dynamic are as follows:

$$\frac{D\mathbf{v}_a}{Dt} = \sum_b m_{0b} \left(\frac{\mathbf{P}_a \mathbf{L}_{0a}^{-1}}{\rho_{0a}^2} + \frac{\mathbf{P}_b \mathbf{L}_{0b}^{-1}}{\rho_{0b}^2}\right) \bullet \nabla_a W_{ab} + \frac{\mathbf{f}_a^{HG}}{m_{0a}} + \mathbf{g}_a \quad (31)$$

where $\mathbf{P}_a$ and $\mathbf{P}_b$ are the first Piola-Kirchhoff stress tensor of particle $a$ and $b$, respectively. $\mathbf{f}_a^{HG}$ is a penalty force used to control the spurious zero-energy modes that arise from the rank deficiency within the SPH approximation. To calculate the first Piola-Kirchhoff stress, the discretized form of deformation gradient should be given:

$$\mathbf{F}_a = \sum_b \frac{m_{0b}}{\rho_{0b}} (\mathbf{x}_b - \mathbf{x}_a) \otimes \mathbf{L}_{0a}^{-1} \nabla_{0a} W_{0ab} \quad (32)$$

where $\mathbf{x}$ is current coordinates of elastic particle. $\mathbf{L}_{0a}^{-1}$ represents the inverse matrix of the correction matrix of the initial configuration. The relevant strain is

$$\varepsilon = \frac{1}{2}(\mathbf{F}^T \mathbf{F} - \mathbf{I}) \quad (33)$$

where $\mathbf{I}$ is identity matrix. For a hyperelastic material, such as polyurethane, the Saint Venant-Krichhoff constitutive model relates the Green-Lagrange strain to the second Piola-Kirchoff stress tensor via:

$$\mathbf{S} = \lambda tr(\varepsilon)\mathbf{I} + 2\mu\varepsilon \quad (34)$$

$$\lambda = \frac{E\nu}{(1+\nu)(1-2\nu)} \quad (35)$$

$$\mu = \frac{E}{2(1+v)} \tag{36}$$

where **E** is the Young's modulus and *v* is the Poisson ratio. Finally, the first Piola-Krichoff stress is

$$\mathbf{P} = \mathbf{FS} \tag{37}$$

$\mathbf{f}_a^{HG}$ is described by a formula from linear spring theory as:

$$\mathbf{f}_a^{HG} = -\frac{1}{2}\alpha \sum_b \frac{V_{0a}V_{0b}W_{0ab}}{X_{ab}^2}\left(E_a\delta_{ab}^a + E_b\delta_{ba}^b\right)\frac{\mathbf{x}_{ab}}{x_{ab}} \tag{38}$$

where α = 0.1 is proved to ensure satisfactory results. To account for the emergence of spurious zero-energy modes, a corrective force derived from the hourglass control method of reduced-integration finite elements is introduced. A separation error vector between two particles is defined and projected onto the separation vector.

$$\delta_{ab}^a = \frac{(F_a\mathbf{X}_{ab} - \mathbf{x}_{ab}) \bullet \mathbf{x}_{ab}}{x_{ab}} \tag{39}$$

where $X_{ab}$ denotes the initial coordinate of elastic structure.

The acceleration of a fluid particle is determined by the contributions of all adjacent fluid, structural, and boundary particles. The dynamic boundary condition method is adopted, more details of wave-structural coupling are shown in O'Connor et al. (2021). In the numerical flume, the relaxation zone is used for wave making and the damping zone is used for wave absorption (Han and Dong, 2025).

**3.3 Validation and dataset construction**

For the interaction between waves and a vertical elastic breakwater, Fig. 5 shows the comparison of interaction between waves and vertical elastic breakwater. It can be seen that the numerical model can reproduce the interaction process well. Figs. 4 and 5 show the results of the elevation of WG1-WG4 and motion response of vertical elastic breakwater. Fig. 4 shows that the SPH simulations reproduce the motion response of the elastic plate under wave action with good fidelity. The predicted horizontal and vertical displacements agree with the physical results in both phase and overall trend, with only minor differences at certain peaks and in high-frequency components. Fig. 5 further demonstrates that the SPH model captures the wave evolution around the elastic vertical plate accurately. On the incident side (WG1, WG2), the primary waveforms and phases are reproduced well, with WG1 showing the closest agreement, while slight deviations appear at WG2 due to enhanced scattering and nonlinear effects. On the lee side (WG3, WG4), the model reasonably reflects energy attenuation and phase variation, remaining consistent with the experiments except for small discrepancies. Overall, the SPH model proves to be a reliable and accurate tool for simulating the hydrodynamics of elastic structures.

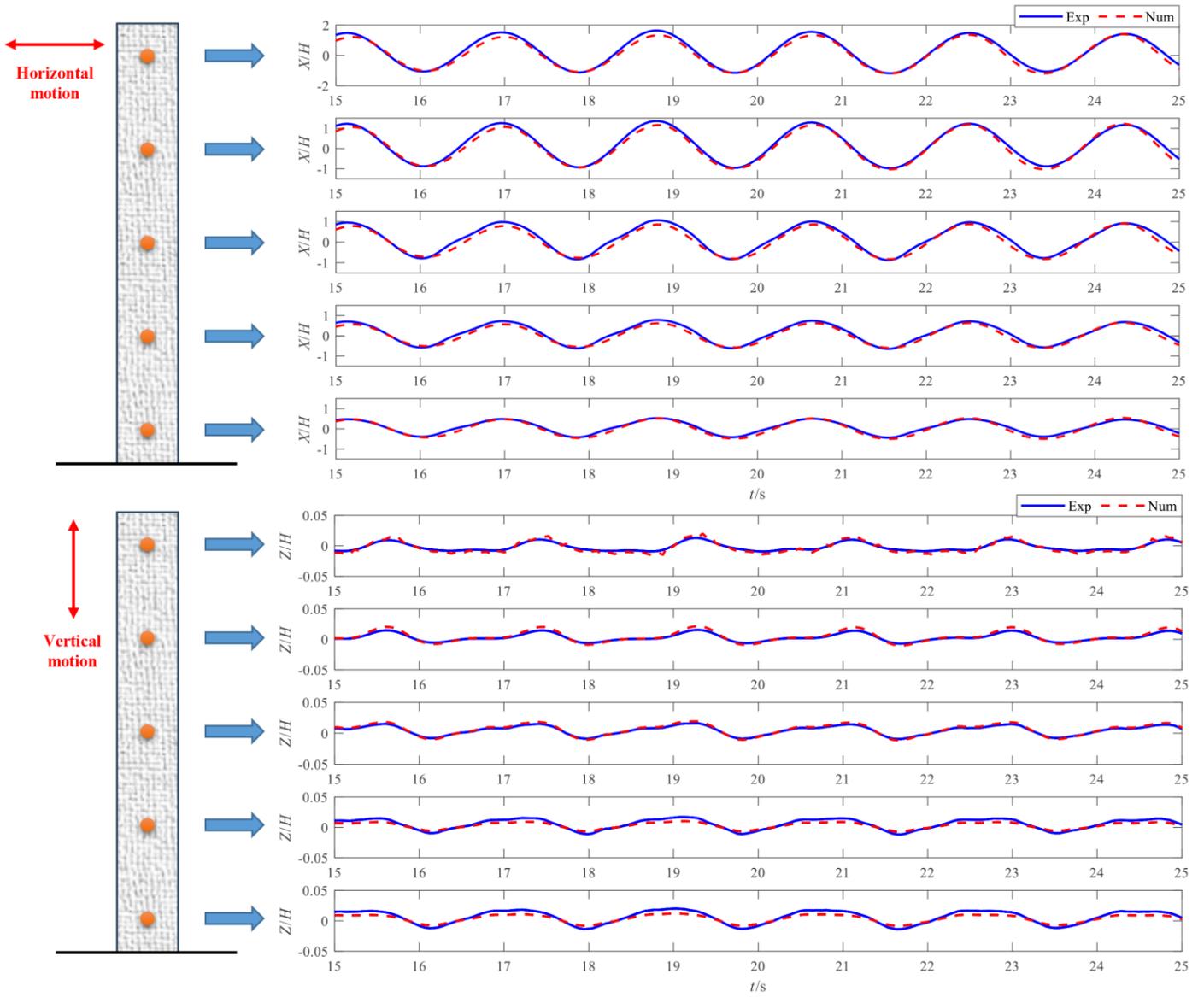

Fig. 4. Horizontal and vertical motion comparison of the physical and numerical results

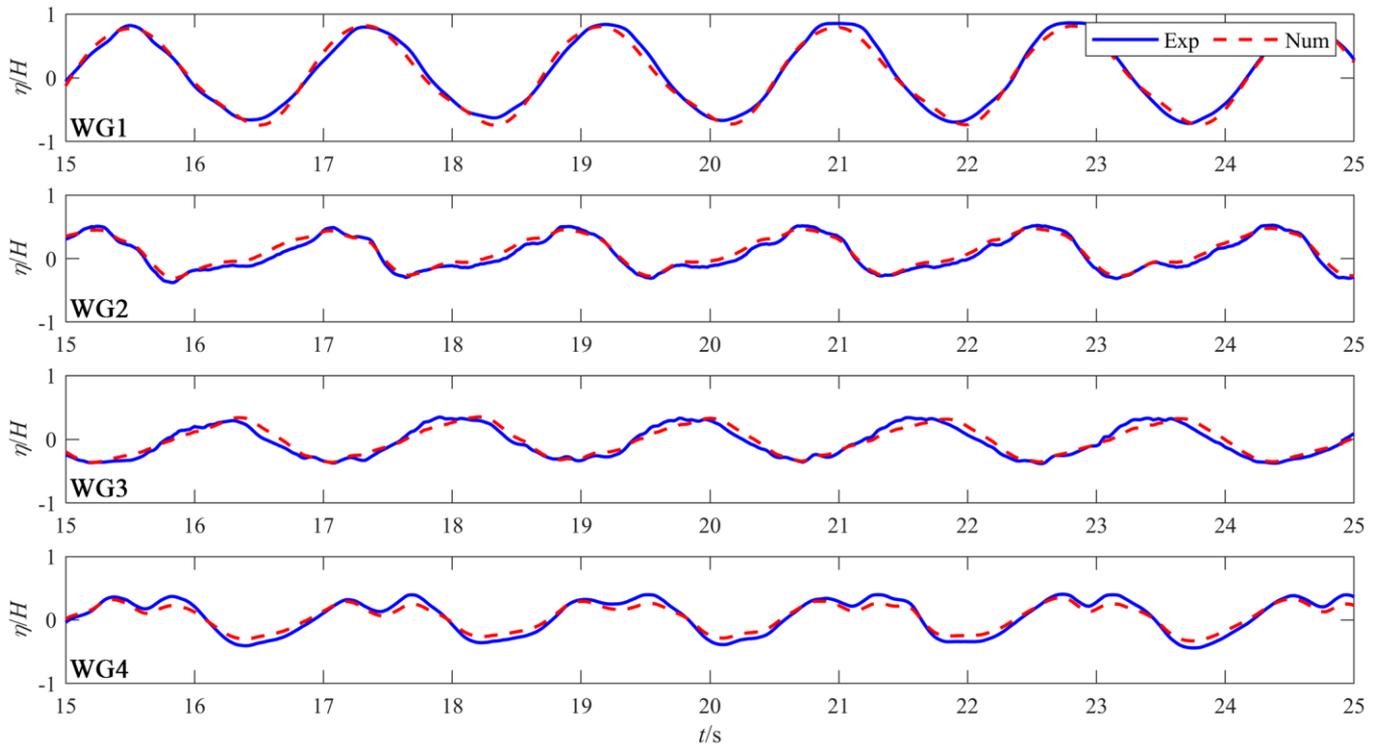

Fig. 5. Wave elevation comparison of the physical and numerical results

In this study, based on the validated numerical model, generates data closely resembling real physical elevation and responses to construct a dataset for predicting the motion response of two-body elastic vertical breakwaters (EVB1, EVB2), as shown in Fig. 6. This approach effectively alleviates the limitation of insufficient physical data. The numerical simulations are conducted under irregular wave conditions that more closely represent realistic sea states. The JONSWAP spectrum with a peak enhancement factor 3.3 is employed (Han et al., 2021). The wave conditions are listed in Table 1, and the wave making time was 500 s. The structure and material parameters of each elastic vertical plate are the same as those mentioned above. The measured quantities consist of free-surface elevations at 12 wave gauges (WG1–WG12) and the displacements of the elastic plate top in the X- and Z-directions.

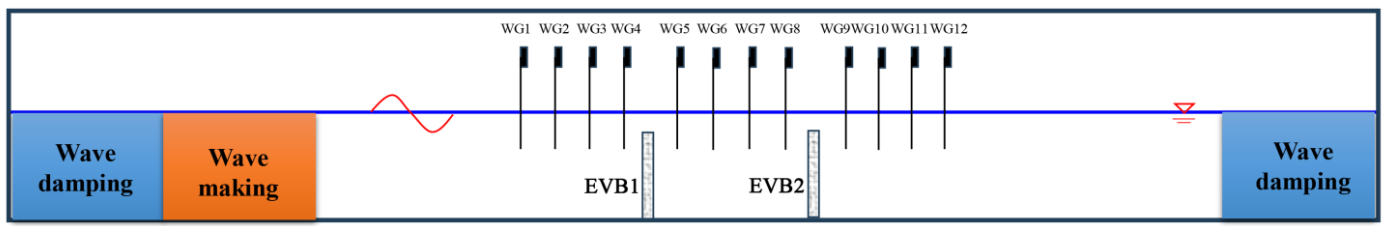

Fig. 6. The numerical model employed for dataset construction

Table 1 Wave conditions

| Water depth | $H_{1\%}$/m | $H_s$/m | $T_m$/s | $T_p$/s |
|---|---|---|---|---|
|  | 0.136 | 0.08 | 2.04 | 2.4 |
|  | 0.198 | 0.12 | 1.36 | 1.6 |
| 0.8 | 0.203 | 0.12 | 1.70 | 2.0 |
|  | 0.204 | 0.12 | 2.04 | 2.4 |
|  | 0.204 | 0.12 | 2.38 | 2.8 |
|  | 0.273 | 0.16 | 2.04 | 2.4 |

## 4. Experimental Setup

To evaluate the proposed model, we employ four metrics that assess prediction accuracy: *MAE*, root mean square error (*RMSE*), spectral mean absolute error (*SMAE*) and the coefficient of determination ($R^2$). Among these metrics, *MAE* and *RMSE* quantify the deviation between predicted and measured values. For $R^2$, values approaching 1 indicate excellent model fitting performance. The *SMAE* assesses the fidelity of spectral characteristics between predicted and measured values to evaluate model performance in the frequency domain. They can be calculated using Eqs. (40) ~ (43):

$$MAE = \frac{1}{n}\sum_{i=1}^{n}|y_i - \hat{y}_i| \tag{40}$$

$$RMSE = \sqrt{\frac{1}{n}\sum_{i=1}^{n}(y_i - \hat{y}_i)^2} \tag{41}$$

$$R^2 = 1 - \frac{\sum_{i=1}^{n}(y_i - \hat{y}_i)^2}{\sum_{i=1}^{n}(y_i - \bar{y})^2} \tag{42}$$

$$SMAE = \frac{1}{m}\sum_{k=1}^{m}|\mathcal{F}(y)_k - \mathcal{F}(\hat{y})_k| \tag{43}$$

where $y_i$ represents the measured values, $\hat{y}_i$ denotes the predicted values, $n$ is the total number of samples, $\bar{y}$ and $\bar{\hat{y}}$ are the mean values of measured and predicted values, respectively. $\mathcal{F}(y)_k$ and $\mathcal{F}(\hat{y})_k$ represent the FFT coefficients of the measured and predicted values at frequency bin $k$, respectively, and $m$ is the number of frequency bins.

To comprehensively evaluate the performance of the proposed model, we evaluate the proposed model against baselines of two categories: 1) RNN-based Methods: LSTM (Hochreiter and Schmidhuber, 1997), TCN (Bai et al., 2018), and LSTM-CNN (Zhao et al., 2017). 2) Transformer-based Methods: Transformer (Vaswani et al., 2017), Informer (Zhou et al., 2021), Autoformer (Wu et al., 2021), FEDformer (Zhou et al., 2022), and iTransformer (Liu et al., 2024).

All models were implemented using the PyTorch framework and optimized with the Adam optimizer at an initial learning rate of 1e-4. The learning rate was adjusted dynamically during training using a cosine annealing scheduler. Training was conducted for a maximum of 50 epochs with a mini-batch size of 32. To mitigate overfitting, a dropout rate of 0.1 was applied in feed-forward modules. An early stopping mechanism with a patience of 10 epochs was employed based on validation set performance, and the model with the lowest validation loss was retained for final evaluation. All experiments were performed on a single Nvidia GeForce RTX 3090 GPU.

## 5. Experimental results and analysis

### 5.1 Overall Performance

Table 2 summarises the performance of all models under the wave condition $H_s$ = 0.16 m, $T_p$ = 2.4 s. The proposed model PhysAttnNet delivers the best results across all four evaluation metrics. Best and second-best results are highlighted in bold and underlined, respectively. The proposed model achieves an *MAE* of 0.1862 and an *RMSE* of 0.3218. Compared to the best-performing baseline, these results represent reductions of 16.9% and 10.7%, respectively, demonstrating a substantial improvement in predictive accuracy. It also achieves the highest R² of 0.8446, reflecting the strongest fit to observations. Furthermore, the proposed model achieves the lowest *SMAE* value of 0.9250, indicating superior preservation of spectral characteristics in the frequency domain, further validating its performance advantage. Analysis of baseline models reveals that Transformer-based architectures (e.g., Transformer, Informer) generally outperform RNN-based architectures (e.g., LSTM, TCN), suggesting that Transformer-based architectures possess advantages in predicting motion responses. However, not all Transformer variants guarantee performance improvements. For instance, Autoformer and FEDformer even underperform certain RNN models, suggesting that their architectural designs are mismatched with the motion response, possibly introducing noise or irrelevant dependencies. This phenomenon reflects the limitations of merely increasing model complexity. In contrast, the proposed model effectively addresses this limitation through targeted improvements incorporating physical prior. The results demonstrate that it not only surpasses mainstream Transformer architectures but also circumvents the potential deficiencies of other variants.

Table 2 Overall prediction performance for the case of $H_s$ = 0.16 m, $T_p$ = 2.4 s

| Model type | Model | MAE | RMSE | $R^2$ | SMAE |
| --- | --- | --- | --- | --- | --- |
| RNN-based methods | LSTM | 0.3343 | 0.4908 | 0.6558 | 1.4696 |
|  | TCN | 0.3215 | 0.5005 | 0.7201 | 1.5569 |
|  | LSTM-CNN | 0.3406 | 0.5041 | 0.6445 | 1.5197 |
| Transformer-based methods | Transformer | 0.2340 | 0.3671 | 0.7824 | <u>1.1685</u> |
|  | Informer | <u>0.2240</u> | <u>0.3602</u> | <u>0.8019</u> | 1.3051 |
|  | Autoformer | 0.3542 | 0.5063 | 0.6695 | 1.7302 |
|  | FEDformer | 0.3458 | 0.5165 | 0.6985 | 1.5482 |
|  | iTransformer | 0.3373 | 0.5216 | 0.6147 | 1.4966 |
|  | PhysAttnNet | **0.1862** | **0.3218** | **0.8446** | **0.9250** |

To further evaluate the proposed model, we analyze motion responses along the X- and Z-directions separately. As shown in Fig. 7, PhysAttnNet achieves the best performance on all metrics in both directions of motion. Comparative analysis reveals that prediction errors are higher in the Z direction than in the X direction, indicating that Z-directional motion involves more complex dynamic characteristics, such as higher nonlinearity or coupling effects. Under these conditions, certain baseline models (e.g., FEDformer) experience significant performance degradation, while PhysAttnNet maintains high accuracy, demonstrating superior robustness under dynamically complex conditions.

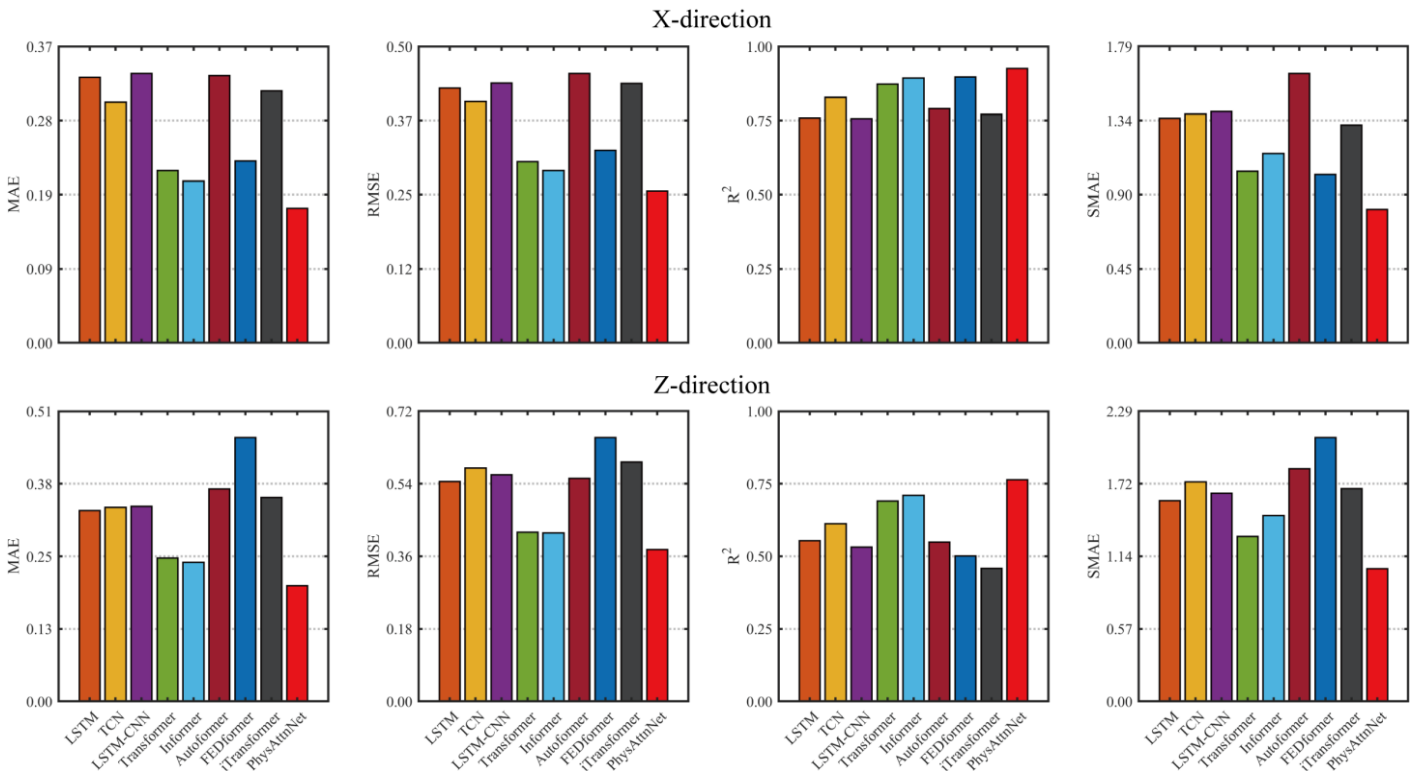

Fig. 7. Quantitative performance comparison on the X- and Z- direction of the motion response.

We evaluate the model predictions by visualizing four motion response variables. Figs. 8 ~ 11 present evaluation results from three perspectives for each plot: A) Time series fitting comparison, B) Peak-valley error analysis, and C) Scatter plot statistical analysis between predicted and measured values. In the time series comparison plots, the orange dashed lines represent the model predictions, while the black solid lines indicate the actual measured data. The results demonstrate that the proposed model achieves excellent agreement between predicted and measured time series across all four response variables, reflecting its superior ability to capture complex motion responses. As shown in Figs. 8 and 10, differences in model performance for X-direction responses are primarily observed in the accuracy of phase and amplitude reproduction. Several advanced Transformer-based models, such as Autoformer and FEDformer, typically exhibit over-smoothing characteristics. While these models adequately capture major trends, they fail to effectively reproduce sharp transitions in response sequences. In contrast, the proposed model achieves exceptional accuracy in waveform reproduction, closely matching the measured responses, including peak values.

The Z-direction exhibits more complex dynamic characteristics compared to the X-direction. As shown in Figs. 9 and 11, the Z-direction response signals contain both transient impulses and a steady, low-amplitude background. This combination poses significant challenges for prediction models. Under these conditions, RNN-based models struggle to identify extreme events. During amplitude spikes, their prediction curves generate only weak perturbations near the baseline, showing significant discrepancies from measured responses in both morphology and magnitude, thus revealing limitations in learning such nonlinear behaviors. While Transformer-based models can identify extreme events, they typically underestimate peak responses. As shown in Fig. 9, the peak values remain considerably lower than measured values. In contrast, the proposed model accurately predicts not only the timing of transient spikes but also their magnitude, with peak values that closely match the measured data. This performance confirms that the physics prior knowledge in PhysAttnNet grants the model the ability to effectively identify extreme events.

Panels B and C in Figs. 8 ~ 11 further validate the above observations from the perspectives of peak-valley error and scatter plot, respectively. In peak-valley error plots, orange and blue bars denote peak and valley prediction errors, respectively. Analysis reveals that the proposed model achieves the lowest peak and valley errors across all response variables. This accuracy in capturing extreme response amplitudes is crucial for reliable assessment of ultimate bearing capacity in ocean engineering applications. The scatter plot analysis (Panel C) demonstrates that the proposed model achieves the highest $R^2$ values and lowest *RMSE* values, confirming its superior prediction accuracy. Other models exhibit marked performance degradation under complex Z-direction conditions, particularly LSTM-based architectures, whose $R^2$ values decline to 0.52–0.64. In summary, the proposed model demonstrates comprehensive advantages in predicting motion responses, considering both qualitative time series morphological fitting and quantitative metrics. The model accurately predicts regular motion responses in the X-direction while effectively capturing extreme and nonlinear responses in the Z-direction, demonstrating the high reliability required for the accurate simulation of real-world dynamic responses.

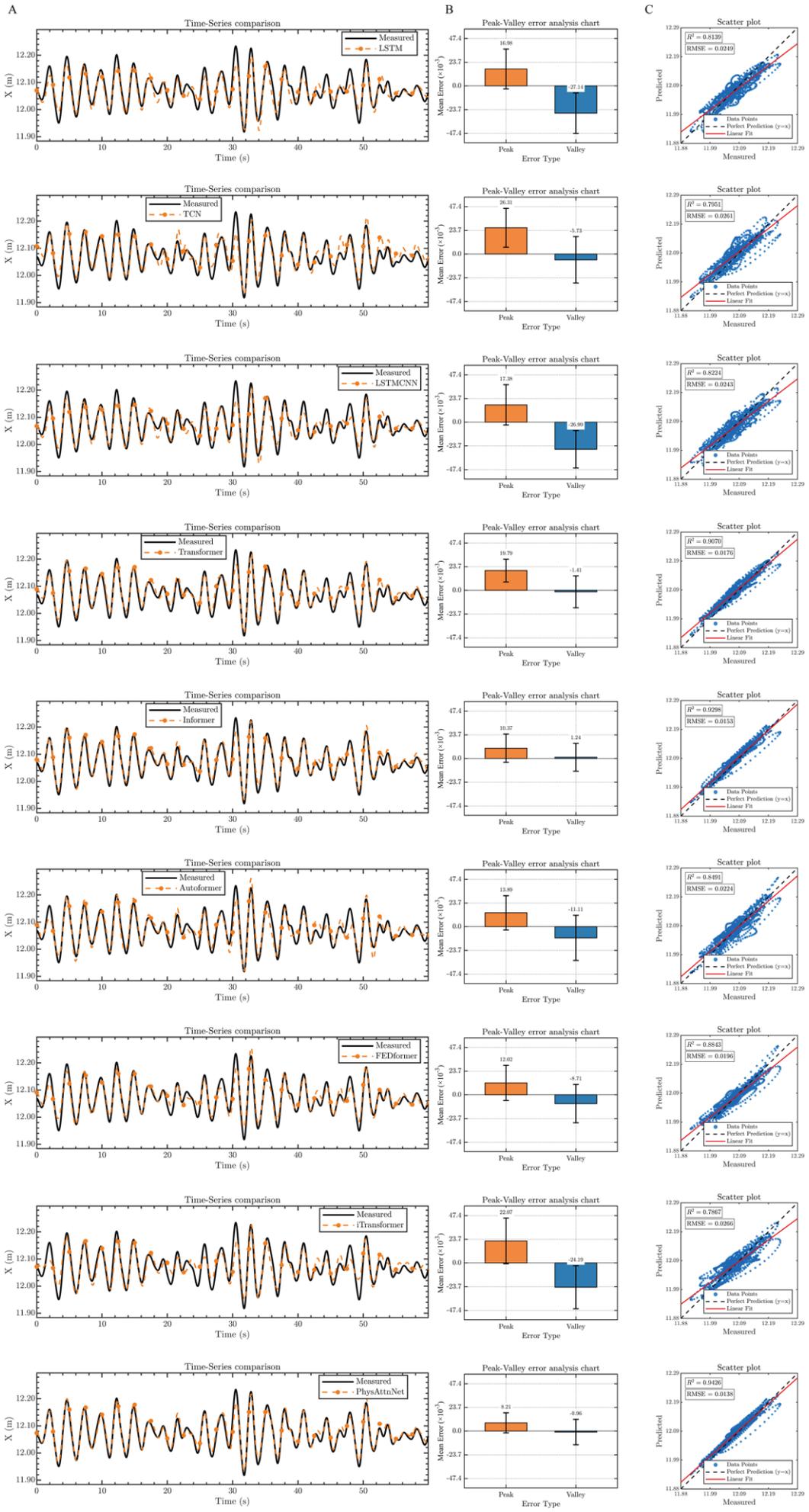

Fig. 8. Comparative evaluation of prediction models for the EVB1 X motion response.

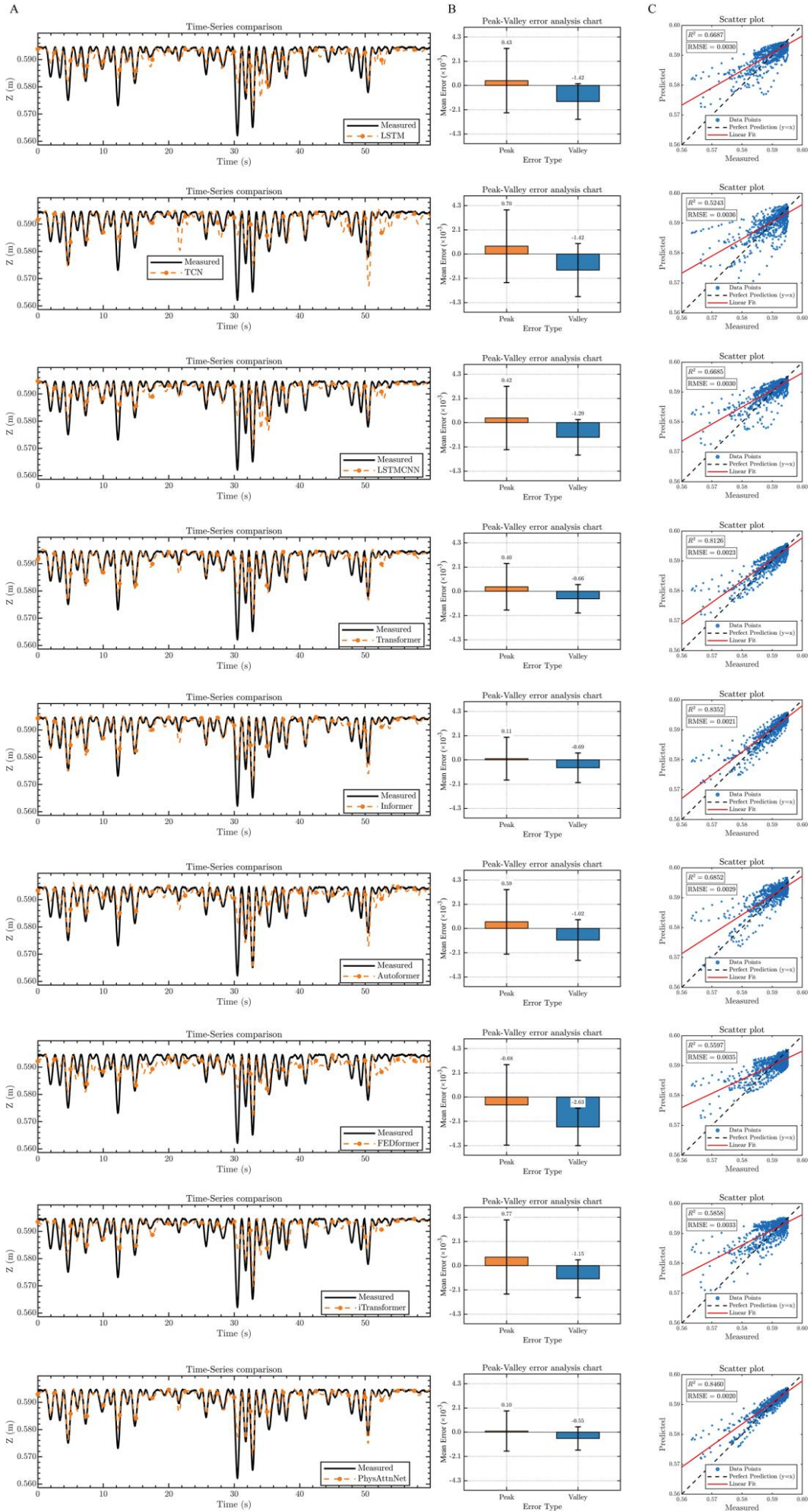

Fig. 9. Comparative evaluation of prediction models for the EVB2 Z motion response.

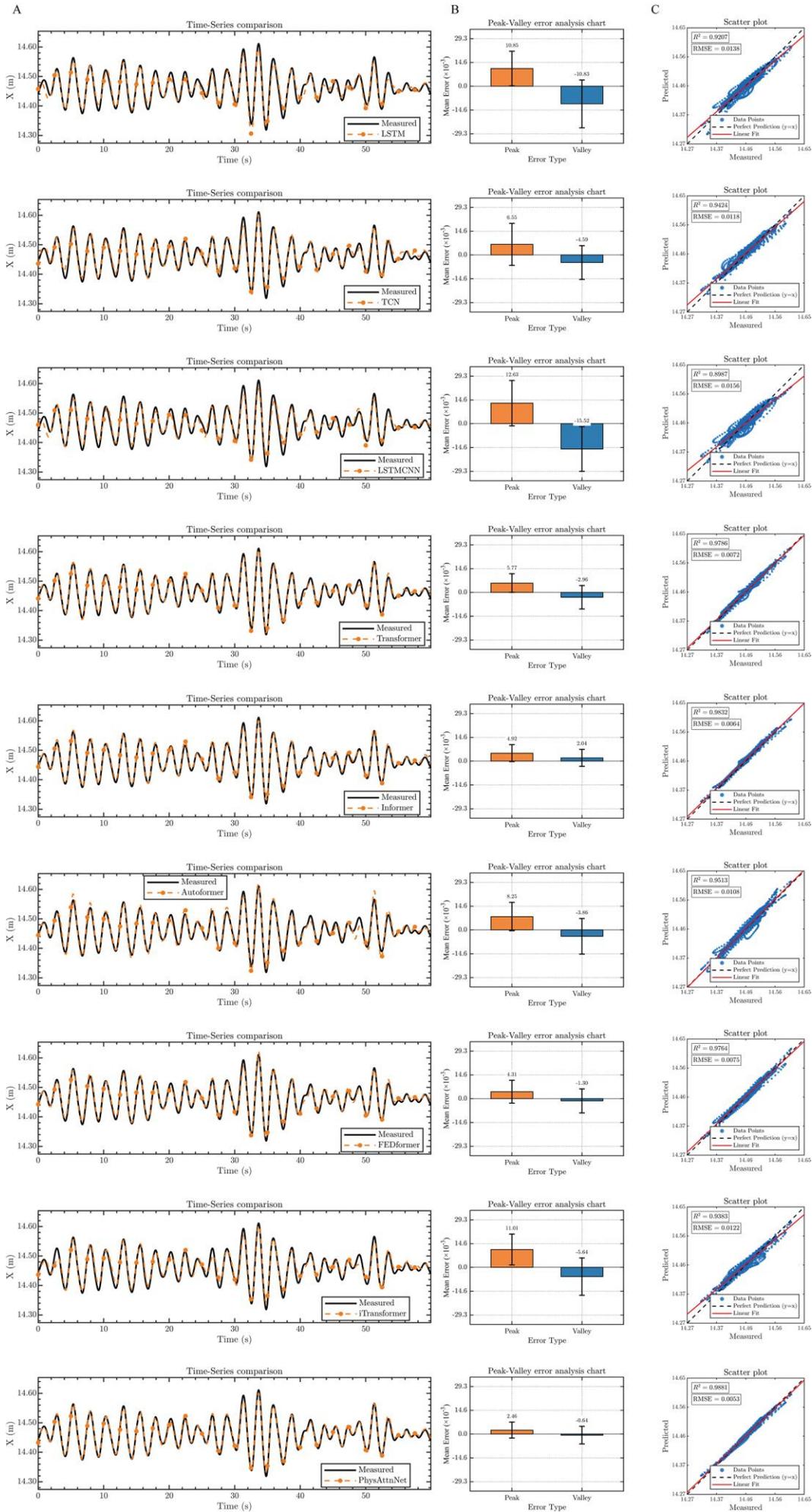

Fig. 10. Comparative evaluation of prediction models for the EVB2 X motion response.

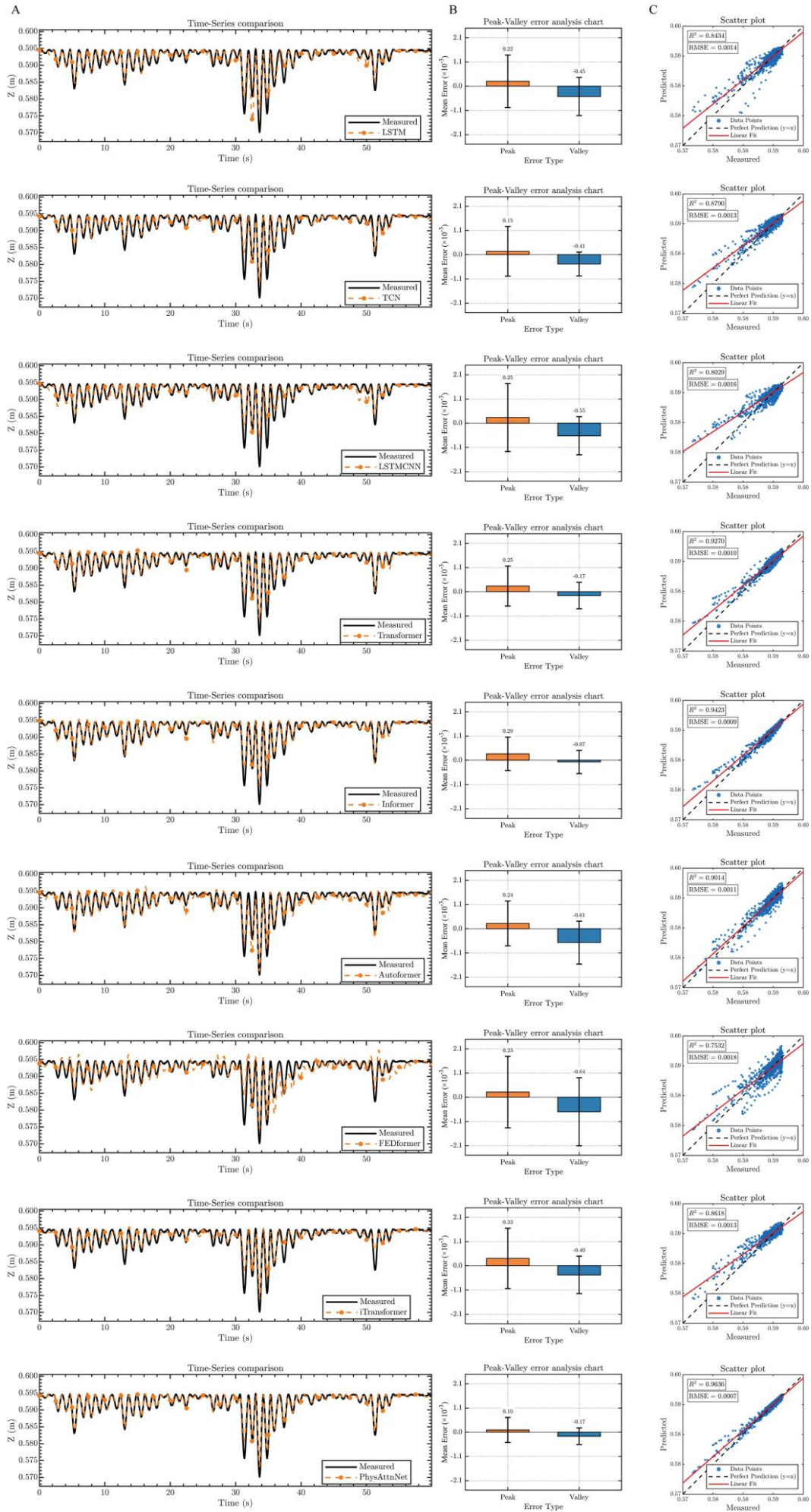

Fig. 11. Comparative evaluation of prediction models for the EVB2 Z motion response.

## 5.2 Performance analysis across varying prediction horizons

To evaluate the performance consistency of the proposed PhysAttnNet across different prediction horizons, we conducted a comprehensive sensitivity analysis over prediction horizons ranging from 6 to 48 time steps. As shown in Fig. 12, all models exhibit a consistent degradation in performance as the prediction horizon increases. Specifically, *MAE*, *RMSE*, and *SMAE* exhibit a progressive increase, while $R^2$ shows a corresponding decline. This trend is characteristic of the error accumulation inherent in iterative time series forecasting. However, the rate of performance decay varies significantly across models. PhysAttnNet consistently outperforms all benchmarks across every tested horizon on all metrics. This advantage is particularly pronounced in the more challenging Z-direction predictions. This stability across varying prediction horizons is attributable to the architectural innovation of the DBSA module. Its embedded temporal decay prior acts as a powerful regularizer, anchoring the model's predictions to the most recent temporal features, thereby preventing the compounding errors that arise from spurious long-range dependencies.

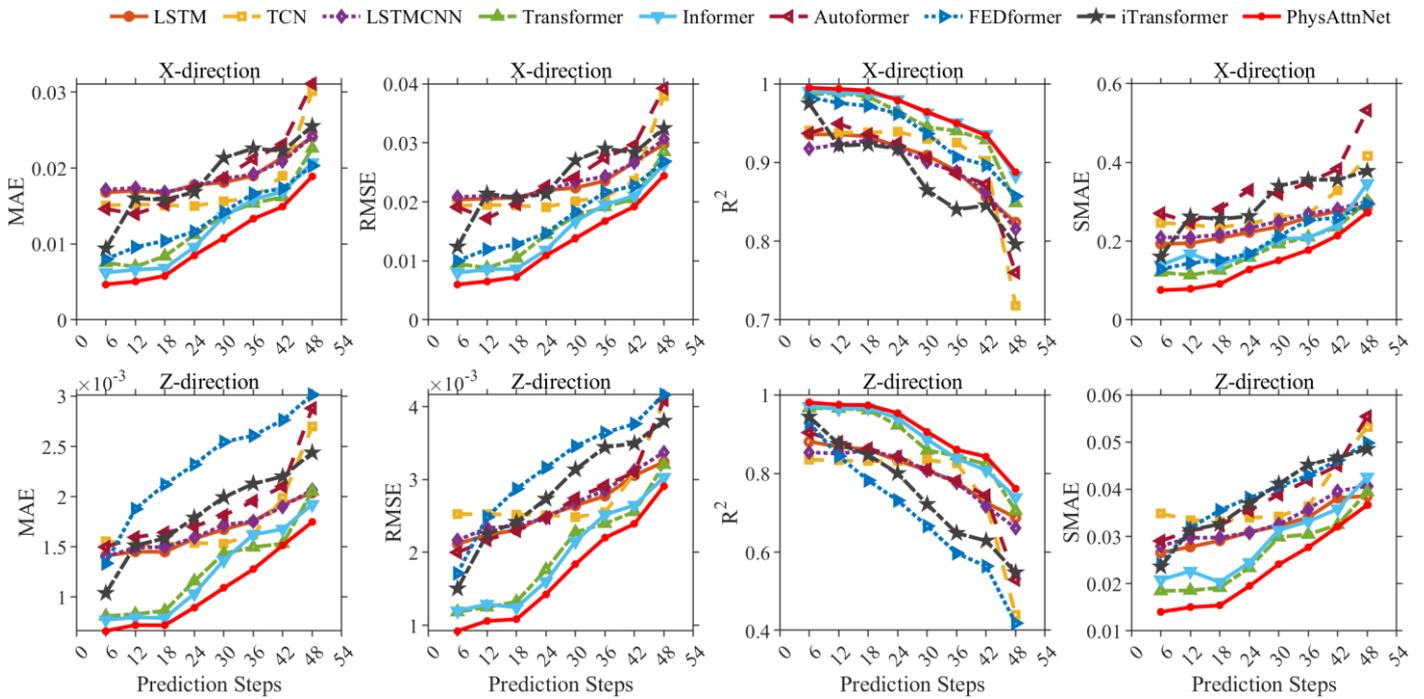

Fig. 12. Performance comparison of different models under various prediction horizons.

To comprehensively evaluate model performance across varying prediction horizons, scatter density plots were constructed to visualize the correspondence between predicted and measured values. Figs. 13 and 14 present these visualization results in a matrix layout for the X and Z directions, respectively, where rows correspond to prediction steps and columns represent distinct prediction models. A consistent trend of performance decay is evident across all models with increasing prediction horizons, manifesting as wider scatter distributions, diminishing $R^2$ scores, and inflating *RMSE*. This observation corroborates the quantitative findings previously presented in Fig. 12. Notably, the proposed model demonstrates superior performance, consistently maintaining the most compact scatter distribution along the ideal diagonal reference line (y=x). Furthermore, it exhibits the most gradual performance degradation across all experimental conditions. Even at the maximum prediction horizon, the model preserves relatively high correlation coefficients and minimal prediction errors. The Z-direction prediction task poses substantially greater challenges than its X-direction counterpart, with all models exhibiting accelerated performance deterioration. At the 48-step prediction horizon, where most benchmark models exhibit severe performance degradation, the proposed model maintains high predictive accuracy with $R^2 = 0.8750$ and *RMSE* = 0.0018.

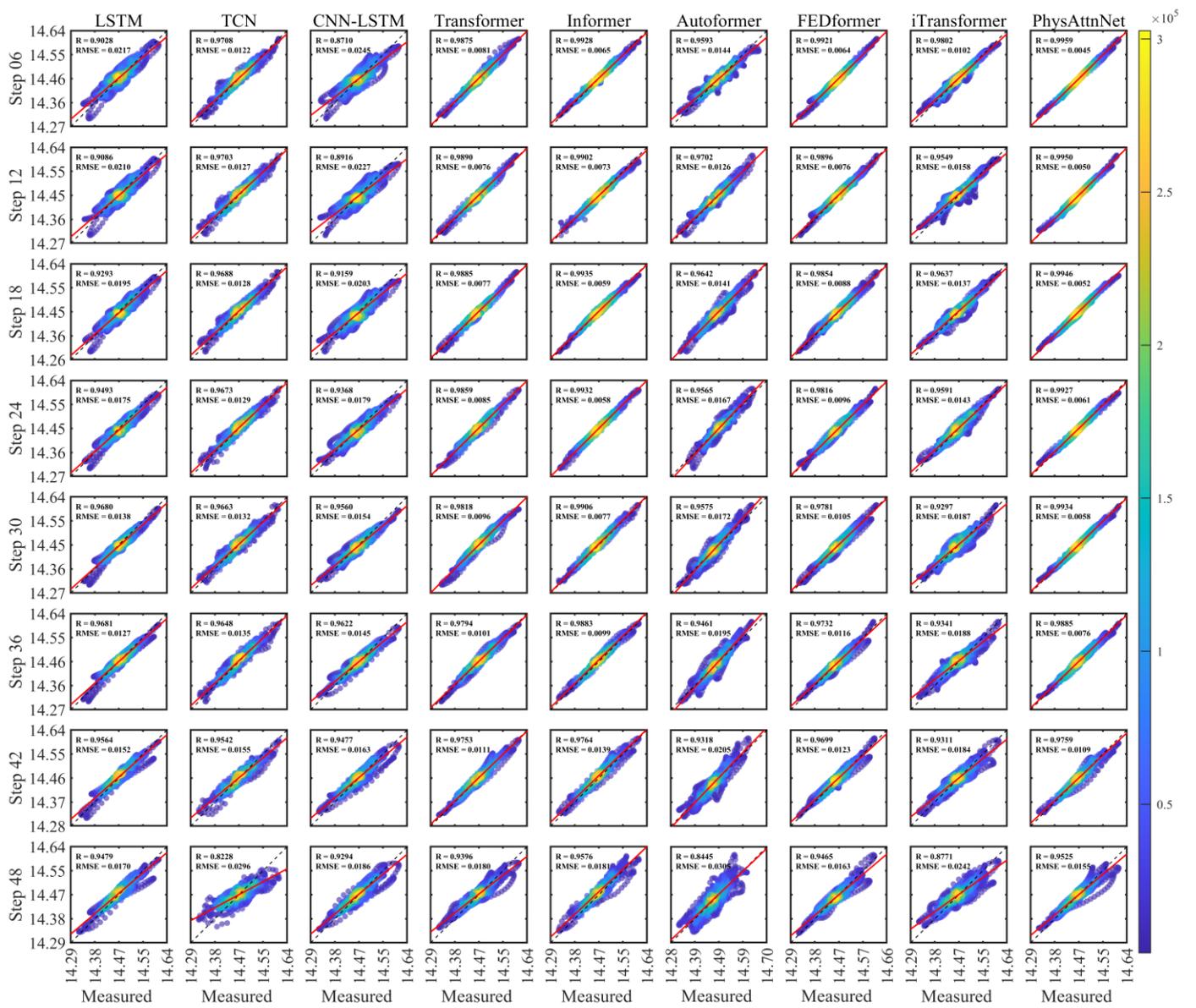

Fig. 13. Scatter density plots for X-direction response at different prediction horizons.

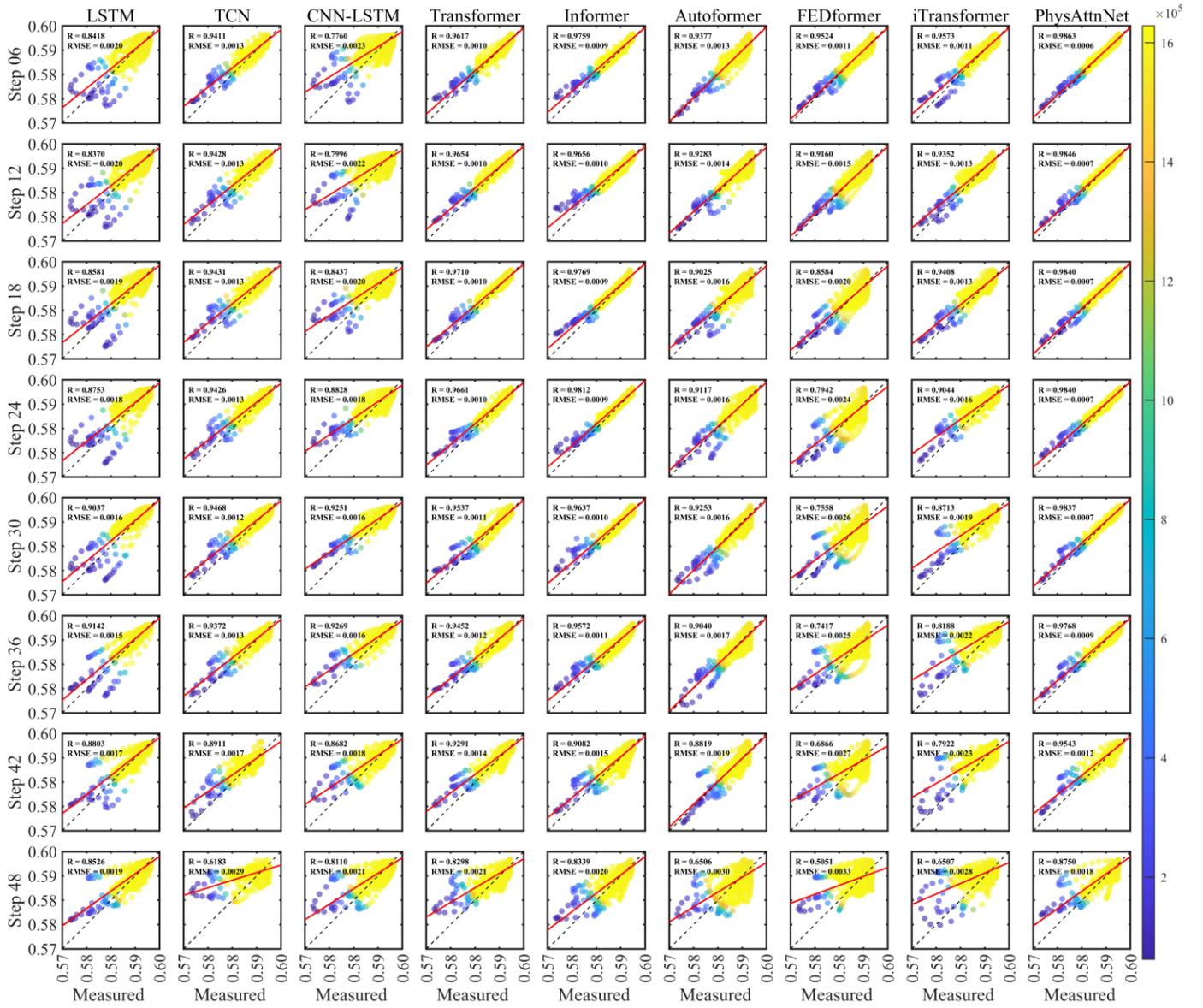

Fig. 14. Scatter density plots for Z-direction response at different prediction horizons.

### 5.3 Ablation study

#### 5.3.1 Effectiveness of model components

This study conducted ablation studies to assess the effectiveness of each core component in the proposed model architecture. Table 3 presents a detailed performance comparison among four ablation variants and the complete model: a) w/o DBSA: replacing DBSA with a standard self-attention mechanism; b) w/o PDG-BCA: replacing PDG-BCA with a standard cross-attention mechanism; c) w/o GCF: employing fixed-weight linear combination; d) w/o time-frequency loss: using only the time-domain loss.

To further elucidate the contribution of the two core attention modules, we analyze the performance impact of w/o DBSA and w/o PDG-BCA. As shown in Table 3, removing either component results in significant performance degradation, confirming their complementary roles in capturing temporal decay and WSI Specifically, the w/o DBSA model exhibits a 9.2% increase in *MAE* and 7.4% rise in *RMSE*, reflecting the critical importance of decay bias in emphasizing recent-history dependencies. Even more critically, removing PDG-BCA causes R² to drop from 0.8446 to 0.8270 and *MAE* to increase by 5.0%, revealing PDG-BCA's indispensable role in characterizing bidirectional coupling.

To validate the effectiveness of the GCF module, an alternative fixed-weight linear combination approach was tested. The w/o GCF shows significant increases of 11.4% and 5.8% in *MAE* and *RMSE*, respectively. This confirms the crucial role of the GCF module in achieving deep interaction between the two types of information by dynamically generating global summaries using PDG-BCA outputs and adaptively integrating them into DBSA.

To validate the effectiveness of the hybrid time-frequency loss function, an alternative approach using only time-domain loss was tested. As shown in Table 3, the w/o time-frequency loss model shows an 8.6% increase in *MAE* and a 25.3% increase in *SMAE* compared to the PhysAttnNet model. The results indicate that the lack of frequency-domain error supervision leads to significant spectral discrepancies, highlighting the core role of the hybrid time-frequency loss function in synergistically optimizing time-domain and frequency-domain accuracy.

Table 3 Quantitative results of the key components of PhysAttnNet.

| Models | Modules | | | | Metrics | | | |
|---|---|---|---|---|---|---|---|---|
| | DBSA | PDG-BCA | GCF | Time-frequency loss | *MAE* | *RMSE* | $R^2$ | *SMAE* |
| w/o DBSA | | √ | √ | √ | 0.2033 | 0.3455 | 0.8386 | 1.0162 |
| w/o PDG-BCA | √ | | √ | √ | 0.1956 | 0.3254 | 0.8270 | 0.9604 |
| w/o GCF | √ | √ | | √ | 0.2074 | 0.3405 | 0.8292 | 0.9978 |
| w/o time-frequency loss | √ | √ | √ | | 0.2022 | 0.3362 | 0.8246 | 1.1592 |
| PhysAttnNet | | | | | **0.1862** | **0.3218** | **0.8446** | **0.9250** |

**5.3.2 Case study**

Fig. 15 presents a comparative visualization analysis between DBSA and standard self-attention mechanisms, encompassing weight matrix heatmaps and time-series prediction results. The heatmap in the upper portion of Fig.15(a) reveals that the standard self-attention mechanism learns dispersed and unstructured weight distributions, with high-weight regions scattered irregularly throughout the matrix. This data-driven attention mechanism violates temporal decay principles. The prediction curves in the lower portion further corroborates this deficiency, exhibiting significant underestimation at initial response peaks and notable phase lag with amplitude attenuation during subsequent oscillation periods. In contrast, DBSA achieves structured reshaping of attention distributions by incorporating temporal decay mechanisms. As illustrated in the heatmap in the upper portion of Fig. 15(b), attention weights are highly concentrated near the main diagonal, forming distinct band structures that align with the physical intuition of "strong-near, weak-far" relationships. When predicting current responses, the model prioritizes information from adjacent time steps while systematically attenuating weights with increasing temporal distance. Consequently, the DBSA achieves substantial improvements in prediction accuracy as shown in the lower portion, predicted curves closely match measured data, precisely capturing peaks and troughs while maintaining excellent phase synchronization and amplitude consistency throughout the entire time domain.

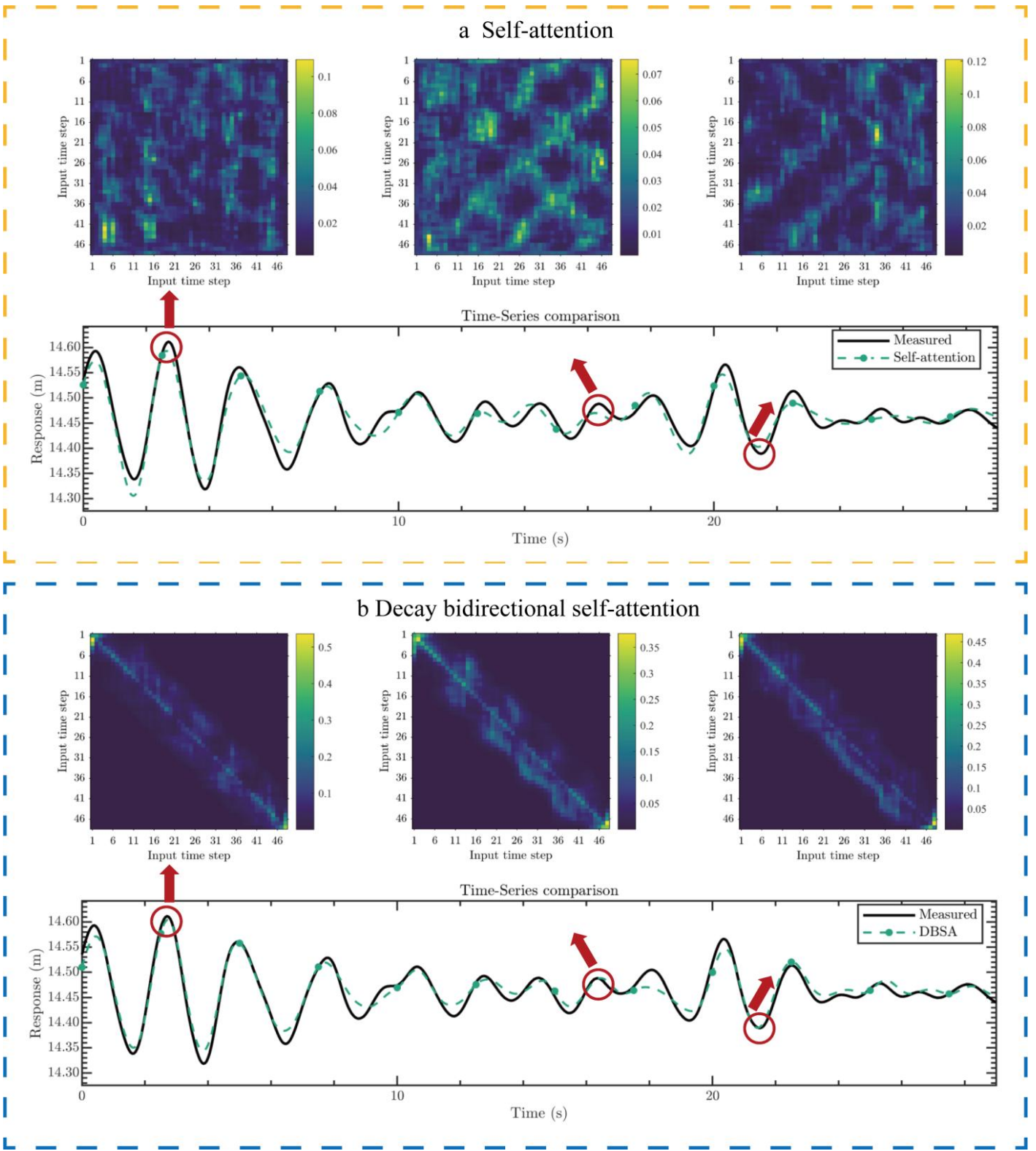

Fig. 15. Comparative analysis of attention score distributions from models with and without the DBSA module.

We visualize the attention weights of the PDG-BCA module to demonstrate its effectiveness in capturing WSI. As shown in Fig. 16(a) and (c), without physics prior guidance, the w/o PDG-BCA model generates diffuse and unstructured attention maps. The attention maps exhibit dispersed focus, assigning high weights across broad temporal intervals without identifying any distinct recurring patterns. This indicates that the w/o PDG-BCA model merely learns statistical correlations from data and fails to identify the inherent phase relationship in WSI. In contrast, the attention maps generated by the PDG-BCA model exhibit highly structured and sparse characteristics, displaying clear periodic diagonal bands, as illustrated in Fig.16(b) and (d). This structured pattern directly reflects the effectiveness of the phase difference guided mechanism. By incorporating phase-biased attention, the PDG-BCA module explicitly guides the model to identify dependencies with phase relationships.

These visualization results confirm that the PDG-BCA module successfully translates the physical concepts of bidirectional coupling and phase relationship into explicit architectural priors.

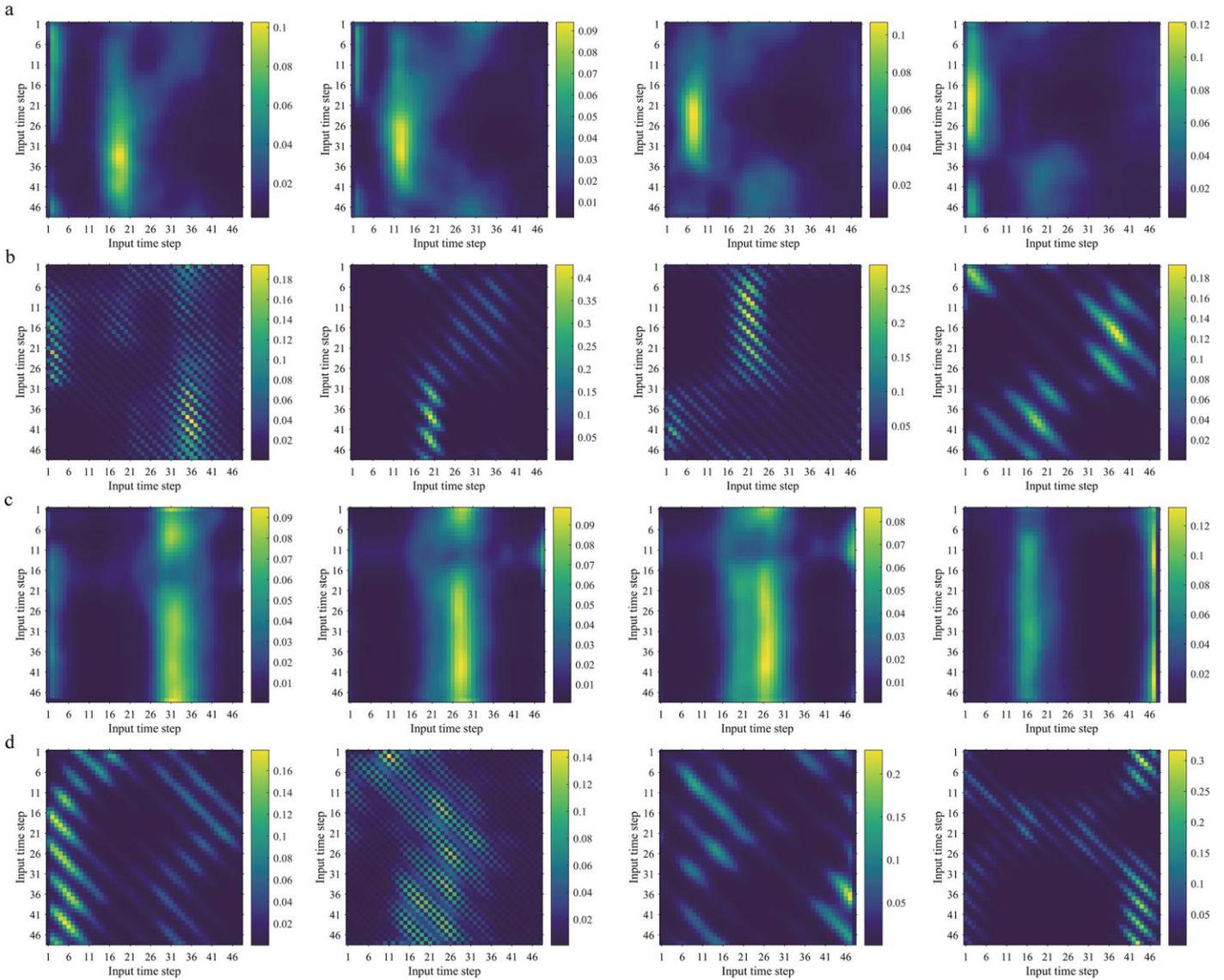

Fig. 16. Comparative analysis of attention score distributions from models with and without the PDG-BCA module.

**5.3.3 Effectiveness of the hybrid time-frequency loss**

We validate the proposed hybrid time-frequency loss by comparing PhysAttnNet variants trained with and without it. Fig. 17 shows CWT analyses of four predictions: (a) and (b) X-direction, (c) and (d) Z-direction responses. CWT visualizes time-frequency energy distributions, enabling quantitative evaluation of frequency fidelity. The top rows of Fig. 17(a) and (b) show that models trained without time-frequency loss approximately capture the dominant 2 - 4 Hz energy band but fail to accurately reconstruct fine spectral details. Their error spectrograms exhibit widespread high-intensity regions near dominant bands, indicating frequency-domain energy leakage and inaccurate spectral reconstruction. In contrast, the bottom rows of Fig. 17(a) and (b) reveal that the hybrid time-frequency loss yields CWT spectrograms closely matching the ground truth, with markedly suppressed error intensities and lower overall error levels. The limitations of without time-frequency loss functions become increasingly pronounced when processing complex Z-direction responses. As demonstrated in the top rows of Fig. 17(c) and (d), extensive high-intensity error regions are observed in error spectrograms for models without time-frequency loss. In contrast, the bottom row results clearly demonstrate the superiority of the proposed loss functions, where the model accurately captures the main energy bands while maintaining only minimal, scattered errors in the corresponding spectrograms. The analysis presented in Fig. 17 provides insight into the significant *SMAE* metric improvements shown in Table 3, which result from the effective suppression of high-intensity error regions in the spectral domain. By imposing explicit frequency-domain constraints, this approach drives the model to generate predictions that achieve both numerical accuracy in the time domain and high spectral fidelity in motion response.

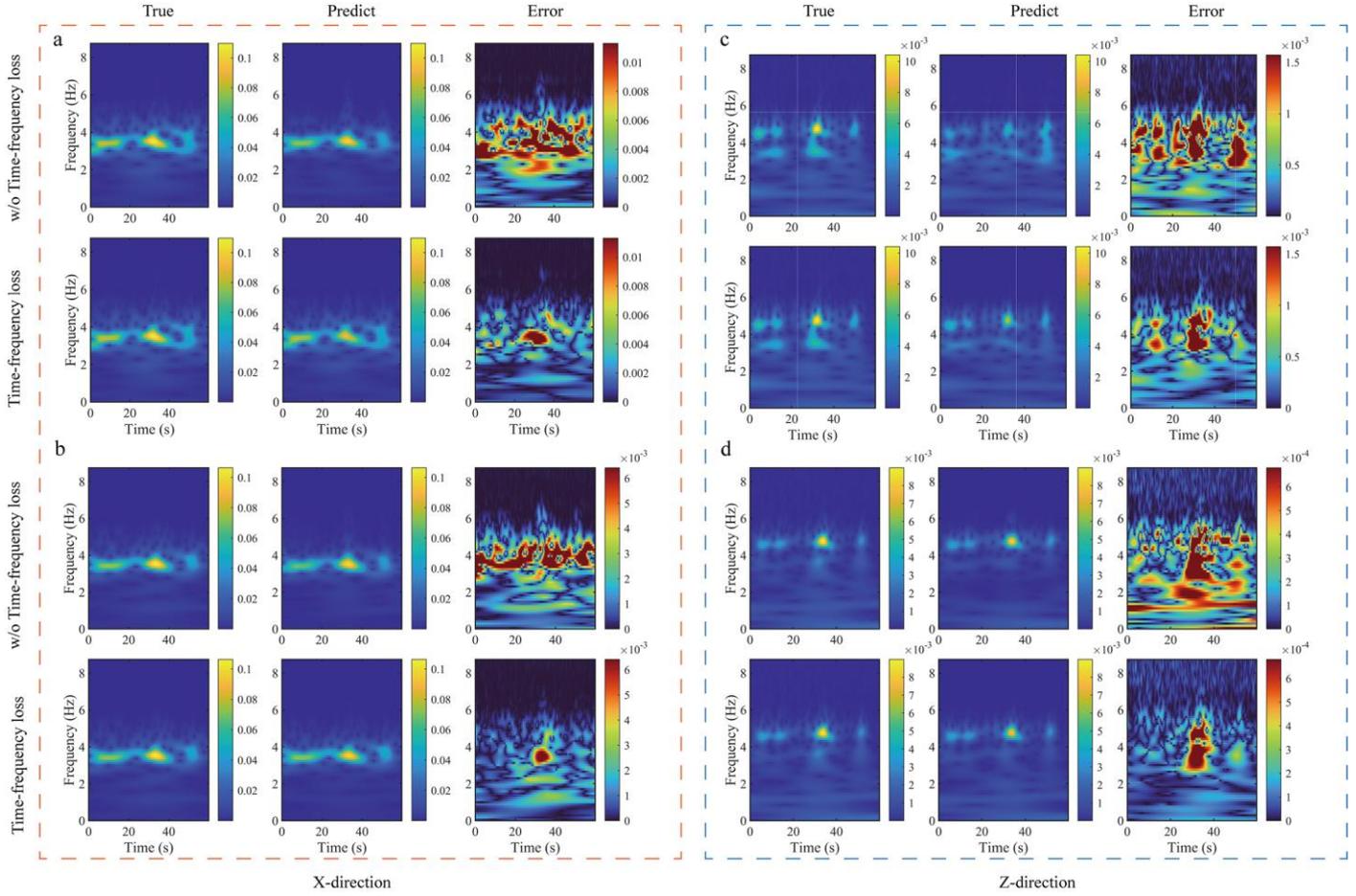

Fig. 17. CWT analysis of prediction results with and without the hybrid time-frequency loss function.

**5.3.4 Sensitivity analysis of key hyperparameters**

To validate the rationality of key hyperparameter selection and evaluate model performance sensitivity, this study conducted a comprehensive analysis of two critical hyperparameters: network depth and the weight coefficient $\lambda$ in the hybrid time-frequency loss function. Network depth determines model representation capability, where shallow networks risk underfitting while excessive depth may cause overfitting and increased computational overhead. Fig. 18 presents the performance evaluation across varying network depths from 1 to 6 layers. The model's performance progressively improved as the number of layers increased to five, demonstrated by a continuous reduction in *RMSE* and a steady rise in $R^2$. This trend, indicating enhanced capability to extract deep-level features, reached its optimum at five layers with an *RMSE* of 0.3218 and an $R^2$ of 0.8447. However, extending beyond this configuration by adding a sixth layer resulted in performance degradation, likely attributable to either overfitting or representational saturation. The weight coefficient $\lambda$, ranging from [0,1], balances time-domain and frequency-domain losses, with $\lambda = 0$ indicates complete omission of frequency-domain constraints. Fig. 19 presents analysis of $\lambda$ impact on model performance. As $\lambda$ increased from 0 to 0.6, both *MAE* and *SMAE* steadily decreased, indicating that moderate frequency-domain constraints improved time-domain prediction accuracy. Optimal comprehensive performance occurred at $\lambda = 0.6$, with minimum *MAE* and *SMAE* values of 0.1862 and 0.9250, respectively. However, performance deteriorated sharply when $\lambda$ exceeded 0.8, indicating that overemphasizing frequency-domain matching disrupts optimization balance. Sensitivity analysis revealed that a 5-layer architecture with a 0.6 weight coefficient achieves an optimal trade-off between predictive accuracy and computational efficiency.

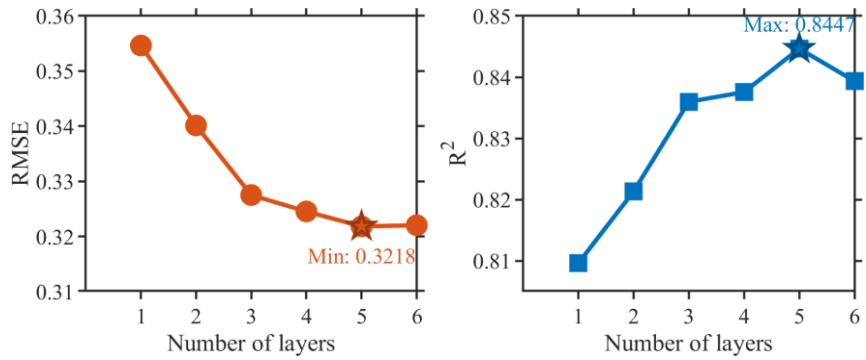

Fig. 18. Impact of network depth on model performance.

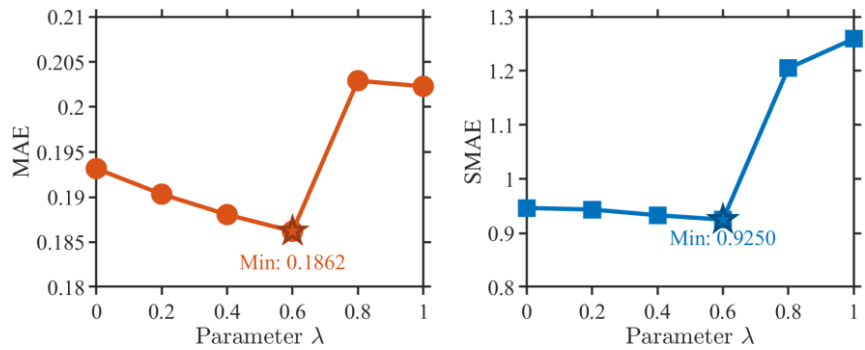

Fig. 19. Effect of the loss weight coefficient on model performance.

## 5.4 Generalization evaluation

To evaluate the generalization capability of PhysAttnNet, this section conducts testing from two aspects: 1) predictive capability for different sea states under zero-shot conditions; 2) performance stability under different material properties. The experiments aim to verify the model's reliability and adaptability under varying conditions.

### 5.4.1. Zero-shot generalization

To evaluate the model's zero-shot generalization capability, we assessed its performance across five datasets representing previously unseen sea conditions. These conditions were defined by distinct combinations of $H_s$ and $T_p$: $H_s = 0.08$ m, $T_p = 2.4$ s; $H_s = 0.12$ m, $T_p = 1.6$ s; $H_s = 0.12$ m, $T_p = 2.0$ s; $H_s = 0.12$ m, $T_p = 2.4$ s; and $H_s = 0.12$ m, $T_p = 2.8$ s. The evaluation was based on the *RMSE* and the *SMAE* to quantify predictive accuracy in the time and frequency domains, respectively.

Fig. 20(a) presents the *RMSE* performance comparison of the proposed model across five distinct sea states. PhysAttnNet achieves the lowest *RMSE* across all sea conditions, demonstrating improvements of 7%-19% over the second-best baseline models, which indicates its capability to consistently capture motion responses regardless of variations in wave parameters. As shown in Fig. 20(b), the *SMAE* is reduced by 16%-32%. This result underscores the model's robust generalization, which is manifested not only in superior numerical accuracy but also in its capacity to predict the frequency spectrum. This advantage stems from the integration of hybrid time-frequency loss and the integration of physical priors, enabling the model to learn generalizable patterns rather than merely fitting the data.

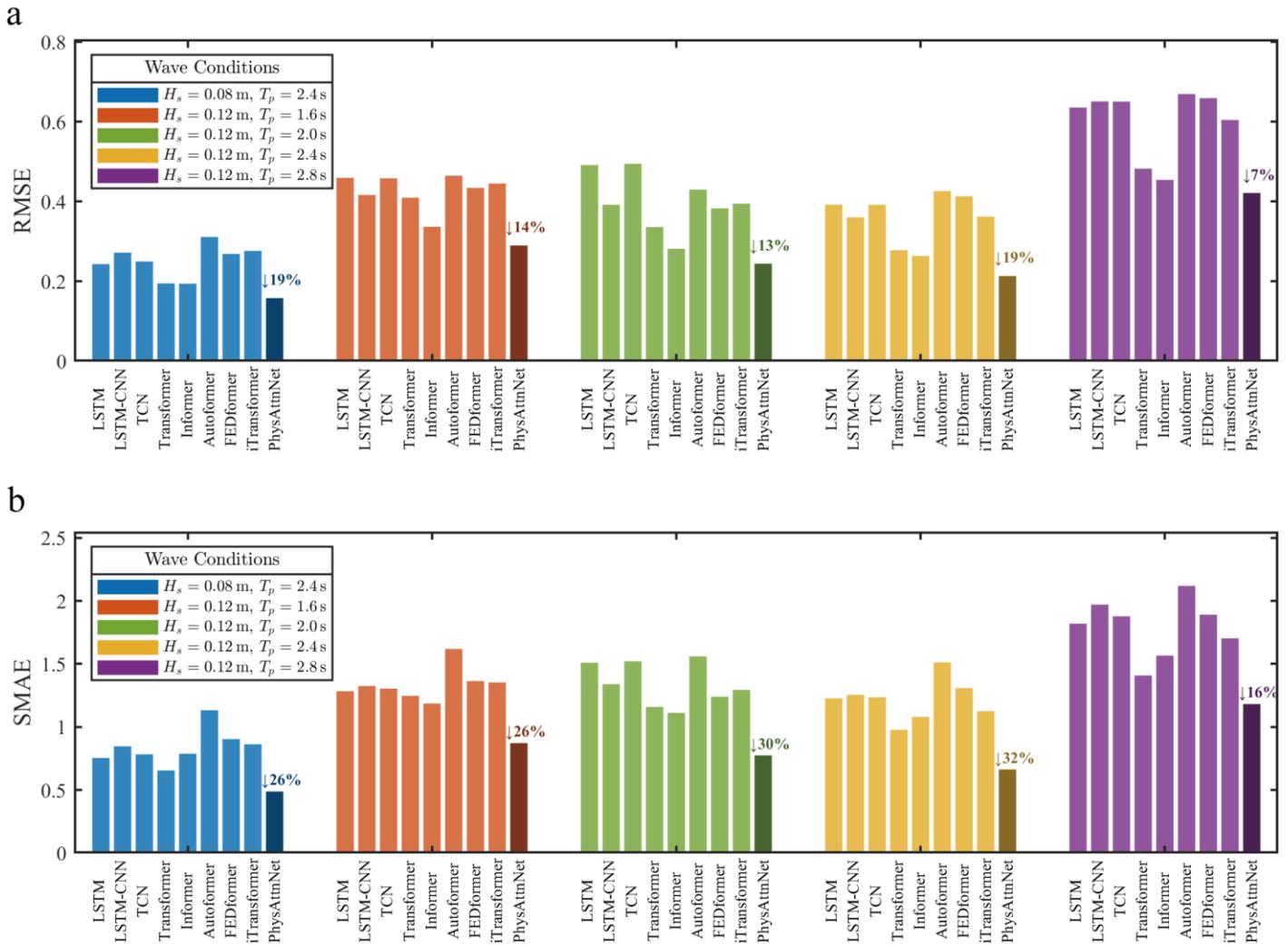

Fig. 20. Generalization performance comparison in terms of *RMSE* and *SMAE* across five diverse sea states.

Fig. 21 presents comparative results between model predictions and ground truth for Z-direction displacement time series on the $H_s$ = 0.12 m, $T_p$ = 2.8 s condition, supplemented with scatter plot analysis. For comparative analysis, we selected the two highest-performing baseline models after PhysAttnNet: Transformer and Informer. As shown in Fig. 21(a) and (b), while both baseline models successfully capture the general oscillatory patterns, they consistently underestimate the amplitude at peaks and troughs. In contrast, Fig. 21(c) demonstrates that PhysAttnNet predictions achieve remarkable concordance with measured data, particularly excelling in capturing extreme values accurately. The scatter plot regressions in Fig. 21(d) - (f) further substantiate these findings. PhysAttnNet achieves an R² value of 0.9507, with its data points tightly clustered around the ideal diagonal line, which indicates that its predictions are both stable and reliable.

In summary, the experimental results validate PhysAttnNet's exceptional generalization performance in diverse sea conditions. The superior generalization capability demonstrated by PhysAttnNet under unknown and dynamically changing oceanic conditions establishes a solid technical foundation for its deployment in practical marine engineering applications.

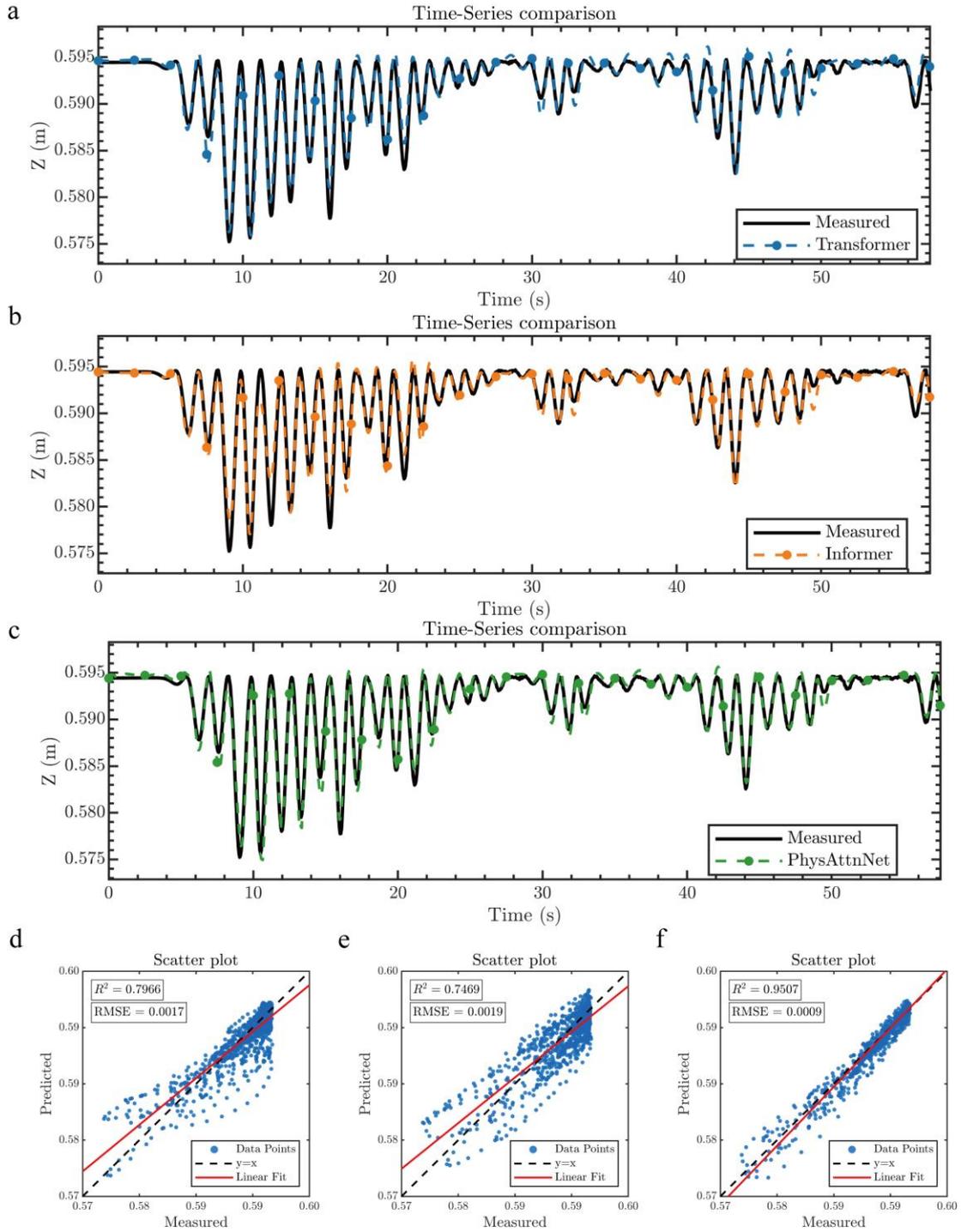

Fig. 21. Predictive performance comparison on the Z-direction displacement time series under $H_s$ = 0.12 m, $T_p$ = 2.8 s.

**5.4.2 Adaptability**

To validate the proposed model's adaptability to variations in material properties, we designed and conducted a series of fine-tuning experiments. Young's Modulus ($E$) was selected as the key control parameter, with quantitative performance comparisons conducted between PhysAttnNet and several baseline models under two conditions: $E$ = 7.75 and $E$ = 10.75. To ensure architectural comparability, all baseline models were restricted to Transformer-based variants, maintaining structural consistency with PhysAttnNet.

As shown in Table 4, PhysAttnNet significantly outperformed all baseline models across all four evaluation metrics under both parameter settings. At $E$ = 7.75, PhysAttnNet achieved $MAE$ = 0.2536, $RMSE$ = 0.3854, $R^2$ = 0.7901, and $SMAE$ = 1.1321, representing 11.4% and 16.1% reductions in $MAE$ and $SAME$, respectively, compared to the second-best model, demonstrating high prediction accuracy and stability under baseline conditions. When $E$ increased to 10.75, despite general performance improvements across all models, PhysAttnNet's relative advantage expanded further, with $MAE$ and $SMAE$ decreasing by 21.8% and 22.3%, respectively, compared to the second-best model, indicating excellent generalization capability under parameter perturbations.

Table 4 Performance metrics of all models under different $E$ conditions.

| Data type | Model | $MAE$ | $RMSE$ | $R^2$ | $SMAE$ |
|---|---|---|---|---|---|
| $E = 7.75$ | Transformer | 0.2861 | 0.4424 | 0.7262 | 1.3498 |
| | Informer | 0.3028 | 0.4536 | 0.7036 | 1.5915 |
| | Autoformer | 0.4203 | 0.5805 | 0.5945 | 1.9555 |
| | FEDformer | 0.3529 | 0.5255 | 0.6650 | 1.6189 |
| | iTransformer | 0.3185 | 0.4895 | 0.6104 | 1.4057 |
| | **PhysAttnNet** | **0.2536** | **0.3854** | **0.7901** | **1.1321** |
| $E = 10.75$ | Transformer | 0.2197 | 0.3023 | 0.7203 | 1.0488 |
| | Informer | 0.1954 | 0.2745 | 0.7401 | 1.1200 |
| | Autoformer | 0.3378 | 0.4343 | 0.6100 | 1.5639 |
| | FEDformer | 0.2848 | 0.3763 | 0.6754 | 1.3263 |
| | iTransformer | 0.2072 | 0.2942 | 0.6867 | 0.9909 |
| | **PhysAttnNet** | **0.1528** | **0.2338** | **0.8072** | **0.7692** |

Fig. 22 presents comparative results between model predictions and ground truth for elastic body Z-direction displacement time series on the $E = 10.75$ dataset, supplemented with scatter plot quantitative analysis. As shown in Fig. 22(f), PhysAttnNet's prediction curve closely aligns with the actual trajectory, with scatter points tightly clustered around the ideal diagonal $y = x$, yielding $R^2 = 0.9542$ and $RMSE = 0.0004$. While Informer performed best among baseline models, PhysAttnNet achieved significant superiority, accurately reproducing not only overall morphology but also precisely capturing amplitude details at peaks and troughs, while maintaining stable fitting even in regions of maximum dynamic variation. The fitted line in the scatter plot is visually indistinguishable from the ideal line, further validating the model's prediction consistency.

In summary, PhysAttnNet explicitly embeds physical priors into the attention mechanism, enabling accurate modeling of both temporal dependencies and underlying system dynamics. This architectural design enables the model to maintain high predictive accuracy under material parameter perturbations, demonstrating superior adaptability compared to approaches relying solely on implicit learning.

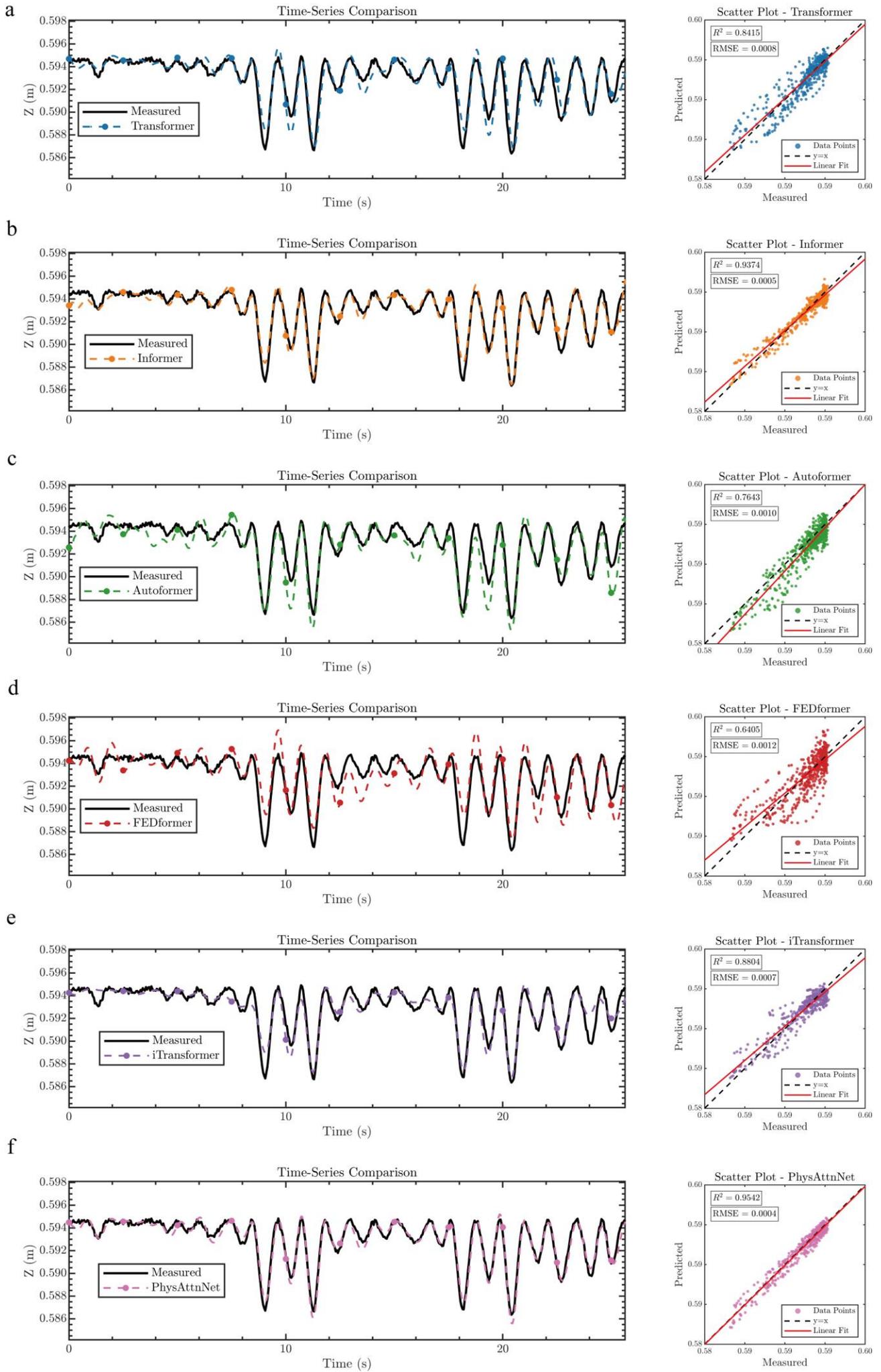

Fig. 22. Predictive performance comparison of the proposed model and baselines on the Z-direction displacement time series.

## 6. Conclusion

This study introduces a novel Physics Prior-Guided Dual-Stream Attention Network (PhysAttnNet) that enhances prediction accuracy and generalization capability for the motion response of elastic Bragg breakwaters. The proposed model addresses the limitations of conventional data-driven models by incorporating physical priors through three key innovations: DBSA, PDG-BCA, and a hybrid time-frequency loss function. The DBSA module incorporates a learnable temporal decay inductive bias, ensuring that motion predictions are dynamically weighted toward recent motion states, thus allowing it to learn the system's inherent decay mechanism. The PDG-BCA module employs a cosine-based bias within a bidirectional cross-computation paradigm to explicitly model the phase relationship and bidirectional interaction between waves and structural motion, while the hybrid time-frequency loss jointly minimizes time-domain prediction errors and frequency-domain spectral discrepancies, ensuring consistency across both domains. By integrating these components, PhysAttnNet facilitates a paradigm shift from superficial data fitting to a physics-prior-guided framework. The main conclusions are summarized as follows:

(1) PhysAttnNet achieves superior prediction accuracy compared to competitive baseline models across all evaluation metrics. Specifically, it reduces *MAE* by 16.9% and *RMSE* by 10.7% relative to the strongest baseline. Time-series comparisons demonstrate its exceptional capability in capturing magnitude and phase of dynamic responses, particularly during extreme kinematic states, such as wave peaks and troughs.

(2) PhysAttnNet demonstrates remarkable generalization and robustness across unseen sea states and structural properties, representing the study's primary contribution. In tests across five diverse, unobserved sea states, the model consistently outperforms benchmarks with *RMSE* improvements of 7%-19% and *SMAE* reductions of 16%-32%. This frequency-domain robustness demonstrates that PhysAttnNet learns generalizable patterns rather than relying on superficial data fitting. Specifically, it internalizes the temporal decay through the DBSA module and captures the dynamics of bidirectional coupling via the PDG-BCA module. This physics prior-guided learning is further reinforced by the hybrid time-frequency loss, which ensures robust generalization across unseen dynamic regimes.

(3) The model's internal mechanisms are both interpretable and demonstrably effective. Ablation studies validate the functional necessity of each component, while attention visualizations offer insight into their learned mechanics. DBSA's attention maps learn a banded-diagonal attention pattern that encodes temporal decay by weighting recent states more heavily. Similarly, the PDG-BCA's periodic stripe patterns capture the signature of phase relationship between waves and structure. These findings demonstrate that the model's performance stems not from data fitting, but from its ability to capture fundamental physical phenomena.

However, this study has limitations that present opportunities for future research. The model's complex architecture presents interpretability challenges, while the dual-stream design's computational requirements increase overhead, potentially affecting real-time applications under resource constraints. Future work should focus on improving computational efficiency and developing advanced interpretability techniques. The model's effectiveness should be validated using broader datasets, including full-scale field measurements and diverse marine structures, to confirm its practical applicability. Additionally, incorporating other environmental factors, such as wind and ocean currents, could enhance the framework's comprehensiveness for ocean engineering applications.

**Acknowledge**

The study was supported by the Natural Science Foundation of Shandong Province (No. ZR2025QC519, No. ZR2023QA090), the China Postdoctoral Science Foundation (Certificate Number: 2024M761844).